\documentclass[journal]{IEEEtran}

\usepackage{graphicx}
\usepackage{subfigure}
\usepackage{threeparttable}
\usepackage{booktabs}
\usepackage{cite} 
\usepackage{bm}
\usepackage{amssymb}
\usepackage{amsmath}
\usepackage{color}
\usepackage{hyperref}

\newtheorem{problem}{\bf Problem}
\newtheorem{remark}{\bf Remark}
\newtheorem{definition}{\bf Definition}
\newtheorem{lemma}{\bf Lemma}

\begin{document}

\title{Encircling General 2-D Boundaries by Mobile Robots with Collision Avoidance: A Vector \\ Field Guided Approach}

\author{
	\vskip 1em
	{Yuan Tian, \emph{Graduate Student Member, IEEE}, Bin Zhang, \emph{Graduate Student Member, IEEE}, Xiaodong Shao, \emph{Member, IEEE}, and David Navarro-Alarcon, \emph{Senior Member, IEEE}
	}
	
	\thanks{
		
		{This work was supported by the Research Grants Council of Hong Kong under grant 15201824. (\textit{Corresponding author: David Navarro-Alarcon})}
		
		{
			Y. Tian, B. Zhang, and D. Navarro-Alarcon are with the Department of Mechanical Engineering, The Hong Kong Polytechnic University, Kowloon, Hong Kong (e-mail: yuan99.tian@connect.polyu.hk; me-bin.zhang@connect.polyu.hk; dnavar@polyu.edu.hk)
			
			X. Shao is with the School of Automation Science and Electrical Engineering, Beihang University, Beijing,
			China (e-mail: xdshao\_sasee@buaa.edu.cn)
		}
	}
}

\markboth{TIAN \MakeLowercase{\textit{et al.}}: Encircling General 2-D Boundaries by Mobile Robots with Collision Avoidance: A Vector Field Guided Approach}{}

\maketitle

\begin{abstract}

The ability to automatically encircle boundaries with mobile robots is crucial for tasks such as border tracking and object enclosing. 
Previous research has primarily focused on regular boundaries, often assuming that their geometric equations are known in advance, which is not often the case in practice. 
In this paper, we investigate a more general case and propose an algorithm that addresses geometric irregularities of boundaries without requiring prior knowledge of their analytical expressions.
To achieve this, we develop a Fourier-based curve fitting method for boundary approximation using sampled points, enabling parametric characterization of general 2-D boundaries. 
This approach allows star-shaped boundaries to be fitted into polar-angle-based parametric curves, while boundaries of other shapes are handled through decomposition.
Then, we design a vector field (VF) to achieve the encirclement of the parameterized boundary, wherein a polar radius error is introduced to measure the robot's ``distance'' to the boundary. 
The controller is finally synthesized using a control barrier function and quadratic programming to mediate some potentially conflicting specifications: boundary encirclement, obstacle avoidance, and limited actuation.
In this manner, the VF-guided reference control not only guides the boundary encircling action, but can also be minimally modified to satisfy obstacle avoidance and input saturation constraints.
Simulations and experiments are presented to verify the performance of our new method, which can be applied to mobile robots to perform practical tasks such as cleaning chemical spills and environment monitoring.
\end{abstract}


\IEEEpeerreviewmaketitle

\section{Introduction}
\begin{figure*}[t]
	\centering
	\subfigure[Chemical spill boundary.]{
		\includegraphics[height=4.0cm]{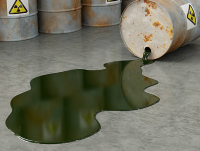}}
	\subfigure[Forest fire boundary.]{
		\includegraphics[height=4.0cm]{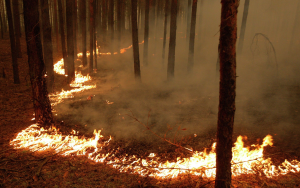}}
	\subfigure[Isoline of observation data \cite{dong2020coordinate}.]{
		\includegraphics[height=4.0cm]{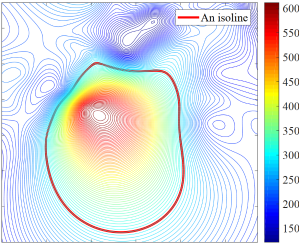}}
	\caption{General boundaries in robot encircling tasks. (a) Encircling the boundary of chemical spills to clean hazardous chemicals. (b) Encircling the boundary of wildfire for fire-fighting. (c) Encircling an isoline of sensory data (e.g., sea temperature, air contaminant concentration) for environment monitoring.}
	\label{intro}
\end{figure*}

\IEEEPARstart{B}{oundary} encircling algorithms are widely applied in robotics for many tasks, such as border tracking and environment monitoring. 
Fig. \ref{intro} shows some potential applications in which a mobile robot may be required to encircle these boundaries to perform tasks, such as cleaning chemical spills, fire-fighting, environmental monitoring, etc.
The term ``boundary encircling'' is also referred to as curve tracking, path following, or contouring control in \cite{michael2017adaptive,yao2021singularity,lam2013model}.
It is a fundamental problem in motion planning and control that aims to steering an autonomous agent to reach and track a closed contour.
The vector field (VF) is a powerful method to generate these types of trajectories, and it has demonstrated good performance across many application scenarios \cite{rezende2018robust,wang2022adaptive,chen2021path}.

Although relevant works have made great progress \cite{goncalves2010vector,kapitanyuk2018guiding,yao2023guiding}, there are still some open challenges in the design of VF-guided boundary encircling methods. 
On the one hand, most existing works require the analytical expression of the boundary (reference path) as a prerequisite.
However, the mathematical expression of the boundary is often unknown in advance and is difficult to determine, particularly for boundaries with geometric irregularities. 
Instead, sampled boundary points are easier to acquire, such as from images or other sensors. 
Consequently, designing efficient VFs for boundaries represented by a set of sampled points is an important problem in practice.
On the other hand, boundary encircling algorithms must be designed to satisfy some motion constraints, such as collision avoidance and input saturation. 
The former requires encircling the boundary while avoiding obstacles along the desired path. 
The latter refers to the conflict between fast/accurate tracking performance and limited linear/angular velocities.
Therefore, the control algorithm should be developed to mediate these potentially conflicting specifications. 
Motivated by the above two concerns, this paper aims to design a VF-guided controller to achieve the encirclement of general boundaries without knowing their analytical expressions in advance, while satisfying collision avoidance and input constraints.

The first step for such an algorithm lies in designing a VF-based guidance law for a general boundary, namely, the desired path. 
To achieve this, distance to the path and direction along the route are two necessary terms to be calculated. 
They are both computed based on the representation of the desired path, for which implicit functions \cite{kapitanyuk2018guiding,goncalves2010vector} and parametric equations \cite{yao2023guiding,liang2015vector,wu2018guidance} are two commonly used representation methods. 
In this paper, we consider the desired path to be represented by a sequence of sampled points, which is often the case for \textit{general} boundaries found in practice. 
A numerical method is presented in \cite{rezende2022constructive}, which can be applied to curves represented by sampled points. 
The minimum distance to the curve is computed by iteratively calculating the distance to these samples, and the tangent direction is estimated between the closest point and its neighbors using Taylor expansion. 
By doing so, the VF can be designed using these discrete points. 
However, it may suffer from computational burden when a large amount of samples are taken into consideration. 
To this end, we present an approximation method to characterize the boundary into a parametric curve, as this analytical representation can help to facilitate the VF design.

The second step is to synthesize a motion control law that considers collision avoidance and limited actuation while encircling the boundary. 
For the former, effective control algorithms include artificial potential fields \cite{lv2023collision}, navigation function methods \cite{chu2022feedback}, explicit reference governor \cite{hu2023spacecraft}, among others. 
These approaches usually require a specific target position to generate a collision-free trajectory, which is not automatically compatible with our boundary encircling task which specifies a desired path. 
To this end, Yao et al. \cite{yao2022guiding} designed a composite VF for following occluded paths, which can achieve path tracking and collision avoidance simultaneously. 
However, input constraints were not involved in this research.
Model predictive control \cite{lee2024nonlinear} can address both obstacle avoidance and input constraints, and it usually follows a hierarchical architecture of upper-level path planner and lower-level tracking controller. 
The control barrier function (CBF) method \cite{zheng2019toward} offers another perspective to cope with system constraints. 
It constructs a forward-invariant set regarding the inequality constraints and then obtains a feasible control command by solving a quadratic programming (QP) problem \cite{ames2017control}. 
This method does not require a high-level planner to generate a reference path. 
Instead, CBF can be regarded as a safety filter that adjusts the reference input to ensure satisfying the constraints. 
For example, Yang et al. \cite{yang2024safety} proposed a safety-critical control allocation scheme using the CBF approach for quadrotor aerial photography tasks to avoid obstacles. 
Chen et al. \cite{chen2024deep} converted system constraints to the constraints on vehicle inputs and presented a safety command governor for autonomous vehicles.

Based on the aforementioned discussions, this paper aims to address the following technical challenges: 
1) characterizing 2-D boundaries using sampled points;
2) encircling the parameterized boundary; and
3) handling motion constraints during the encirclement process.
To address these challenges, we propose a VF-guided reference controller to encircle general 2-D boundaries parameterized by truncated Fourier series. 
Additionally, a synthesized controller is derived using the CBF method to handle obstacle avoidance and input constraints. 
The main contributions of this work are as follows:

\begin{enumerate}
    \item A Fourier-based approximation method for characterizing general boundaries as parametric curves. This method eliminates the need for a predefined boundary equation; instead, it utilizes sampled points to enable the characterization of more general boundaries, including those with irregular or complex shapes.

    \item A VF for guiding the encirclement of the parameterized boundary. The VF serves as a reference controller for mobile robots, enabling to encircle boundaries at a constant speed while maintaining the orientation along the tangent direction of the path. 
    To measure the `distance' from current position to the boundary, we introduce a polar radius error, thus, avoiding the need to solve optimization problems for determining the minimum distance.

    \item A synthesized controller to cope with motion constraints during encirclement. 
    By solving a QP problem which incorporates the CBF method, the VF-guided reference control is minimally adjusted, providing the synthesized controller with the ability to handle constraints. 
    This approach directly corrects the reference controller, rather than the reference path, eliminating the need for a high-level planner and reducing the computational burden.
\end{enumerate}

The rest of this paper is organized as follows: Sec. II presents the mathematical preliminaries; Sec. III describes the proposed methodology; Sec. IV presents the conducted simulations and experiments; Sec. V concludes this article.

\section{Preliminaries}

\subsection{Kinematics}
\begin{figure}[t]
    \centering
    \subfigure[Robot coordinate.]{
    	\includegraphics[height=4cm]{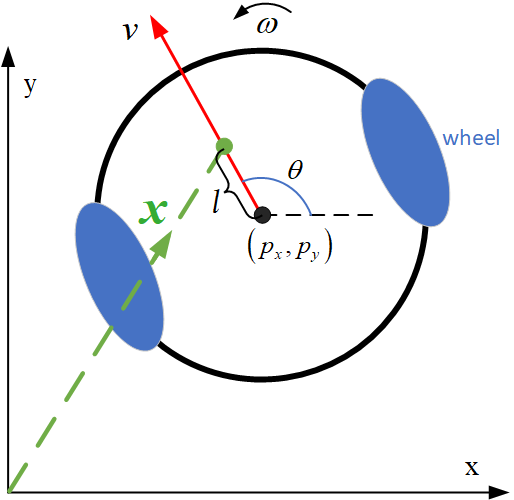}
    	\label{model}}
    \hspace{1em}
    \subfigure[Velocity constraints.]{
    	\includegraphics[height=4cm]{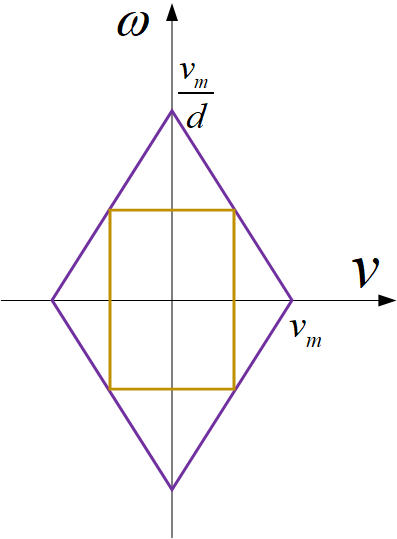}
    	\label{constraint}}
    \caption{Illustration of nonholonomic mobile robot and velocity constraint.}
\end{figure}

A two-wheel differential driving mobile robot is adopted in this paper, and its kinematics is described as 
\begin{equation}
    \dot{p}_x=v\cos{\theta},\ \dot{p}_y=v\sin{\theta},\ \dot{\theta}=\omega
\end{equation}
where $p_x$, $p_y$ are the $x$, $y$ coordinates of the center of the wheeled mobile robot, and $\theta$ denotes the orientation. Define $\bar{\bm{u}}=[v,\omega]^\top\in\mathbb{R}^2$, with $v$ and $\omega$ being linear and angular velocities, respectively. 
We introduce a new coordinate as \cite{siciliano2009robotics}
\begin{equation}
    x=p_x+l\cos{\theta},\ y=p_y+l\sin{\theta}
\end{equation}
where $\bm{x}=[x,y]^\top\in\mathbb{R}^2$ relates to the off-axis point with distance $l$ (arbitrarily small) to the center, as shown in Fig. \ref{model}. With this transformation, the robot is governed by
\begin{equation}\label{kinematics}
    \dot{\bm{x}}=\bm{u}
\end{equation}
with 
\begin{equation}
    \bm{u}=\bm{R}(\theta)\bar{\bm{u}},\ \bm{R}(\theta)=
    \left[\begin{array}{cc}
       \cos{\theta} & -l\sin{\theta}   \\
       \sin{\theta} & l\cos{\theta}  
    \end{array}\right]
\end{equation}
This is a single-integrator kinematics that enables us to freely control the robot position regardless of nonholonomic constraints, and thus facilitates further design and analysis.

\subsection{Obstacle Avoidance and Input Constraints}

In this paper, we are considering static obstacles and assume that each obstacle is enveloped by an enclosing circle with radius $\bar{r}_i$ ($i=1,2,\cdots,n$), with $n$ being the number of obstacles. For ease of design, we regard the mobile robot as a point and transfer its volume to obstacles. In this way, the obstacle avoidance constraint can be formulated as
\begin{equation}\label{state}
    \|\bm{x}-\bm{x}_o^i\|\geq r_o^i,\ i=1,2,\cdots,n
\end{equation}
where $r_o^i=\bar{r}_i+r_b$, $r_b$ is the radius of the smallest enclosing circle of the mobile robot, and $\bm{x}_o^i\in\mathbb{R}^2$ represents the center position of $i$-th obstacle.

To fully exploit the robot's actuation ability, we impose input constraints on the velocity of two driving wheels, instead of restricting linear and angular velocities. The relationship between them can be written as
\begin{equation}
    v=(v_L+v_R)/2,\ \omega=(v_R-v_L)/(2d)
\end{equation}
where $v_L$, $v_R$ are left and right wheel velocity, and $d$ is half the distance between two wheels. Denote he maximum wheel velocity as $v_m$, i.e., $|v_L|\leq v_m$ and $|v_R|\leq v_m$. Then the input constraints can be formulated as
\begin{equation}\label{input}
    -\bm{b}\leq\bm{A}\bar{\bm{u}}\leq\bm{b}
\end{equation}
with $\bm{b}=[v_m,v_m]^\top$, and $\bm{A}=[1,d;1,-d]$. This relates to the diamond-shaped velocity constraint \cite{chen2014tracking}, as illustrated in purple in Fig. \ref{constraint}. Another type of velocity constraint formulated as $|v|\leq a_1$ and $|\omega|\leq a_2$ corresponds to the rectangle input domain in Fig. \ref{constraint}. It is clear that the diamond-shaped velocity constraint \eqref{input} results in a larger velocity domain, and thus can fully exploit the robot's actuation ability.

\subsection{Problem Statement}\label{sec2.3}
Without loss of generality, two-dimensional boundaries can be represented in the following parametric form 
\begin{equation}\label{contour}
    \bm{c}(\rho)=[c_x(\rho),c_y(\rho)]^\top\in\mathbb{R}^2
\end{equation}
with $\rho$ being the parameter. Note that the analytical expression of $\bm{c}(\rho)$ is unknown in advance, and it is approximated later in Sec. \ref{sec3.1}. 

Considering the differential driving mobile robot described by \eqref{kinematics}, and a general boundary \eqref{contour}, the control problem is formulated as follows:

\vspace{-0.5em}
\textit{
\begin{problem}\label{p1}
    The robot is steered to reach and then track the approximation of the boundary $\bm{c}(\rho)$ with a predefined constant speed, keeping its heading along the tangent direction.
\end{problem}}
\vspace{-0.5em}
\textit{
\begin{problem}\label{p2} 
    If the encircling trajectory violates hard constraints \eqref{state} or \eqref{input}, the robot gives top priority to avoiding obstacles and satisfying input constraint, and then continues encirclement. 
\end{problem}}

\section{Methodology} \label{method}
\begin{figure}[t]
    \centering
    \includegraphics[width=8.5cm]{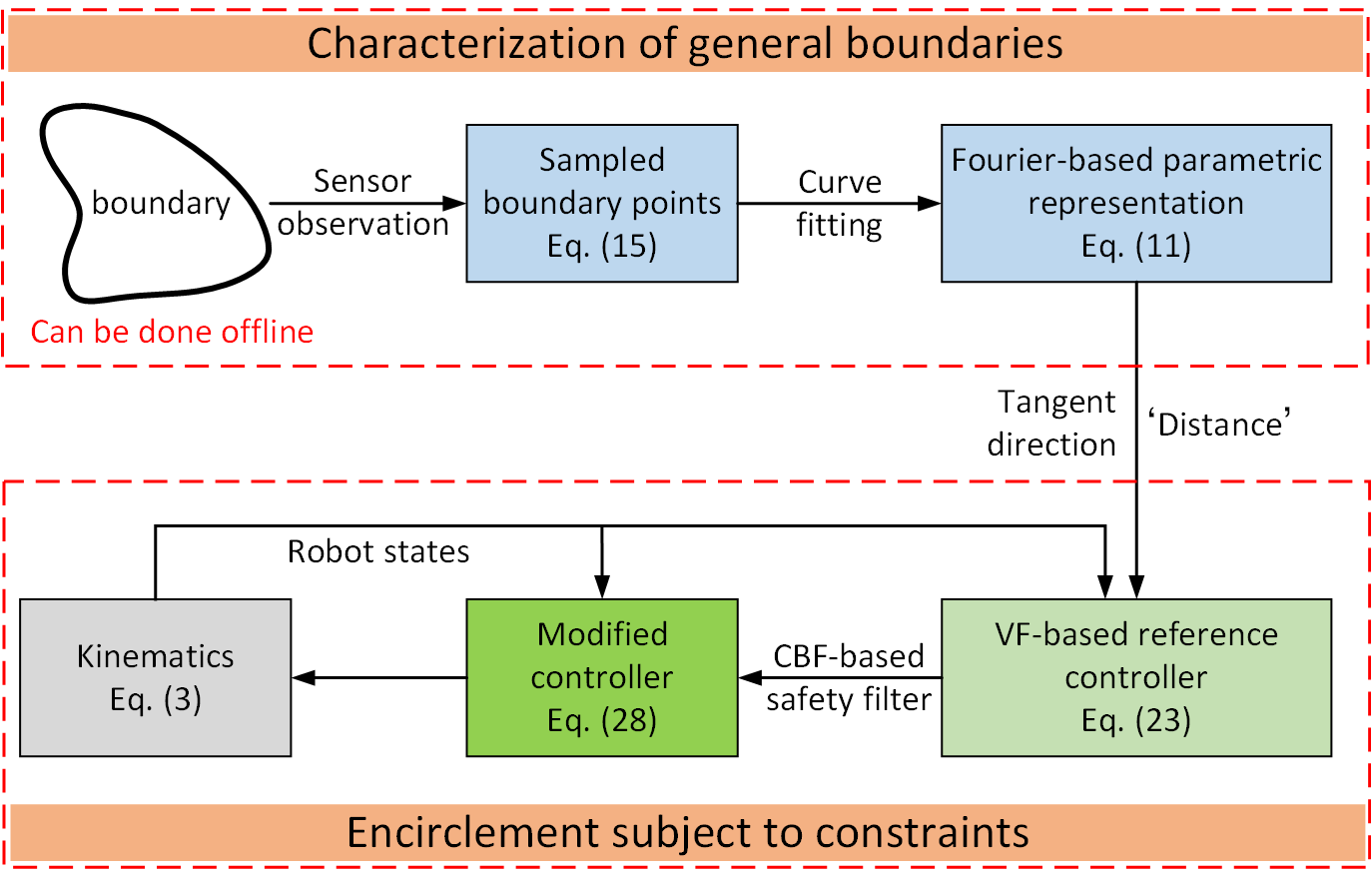}
    \caption{Framework of the proposed algorithm.}
    \label{block}
\end{figure}

This section presents our proposed control methodology. 
We first present a Fourier-based boundary approximation method based on sampled points. The fitted parametric curve is then used to design a VF for boundary encircling. To further address motion constraints, we employ the CBF approach and formulate a QP problem to derive a modified controller. The overall framework of our algorithm is presented in Fig. \ref{block}.

\subsection{Fourier-Based Curve Fitting} \label{sec3.1}

The prerequisite for boundary encirclement is knowing its mathematical expression, which actually defines the reference path to be tracked. Most existing works assume this expression is known in advance and use this accurate boundary model for controller design. However, this is not often the case in practical engineering, and we may need to acquire this expression based on sampled boundary points (easier to obtain from images or other sensors). 
Towards this end, we first need to perform the curve fitting for boundary \eqref{contour}. 
The following definition is given before proceeding.

\vspace{-0.5em}
\textit{
\begin{definition}\label{def1}
    (Star-shaped set)\cite{hansen2020starshaped,rousseas2024reactive}. A set $\mathcal{S}\in\mathbb{R}^n$ is star-shaped if $\exists\ x\in\mathcal{S}$ such that $\forall y\in\mathcal{S}$, the segment $(1-t)x+ty\subset\mathcal{S}$, with $t\in[0,1]$. The point $x$ can be named as a reference point of star-shaped set $\mathcal{S}$.
\end{definition}}
\vspace{0.5em}

\begin{figure}[t]
    \centering
    \includegraphics[width=8.5cm]{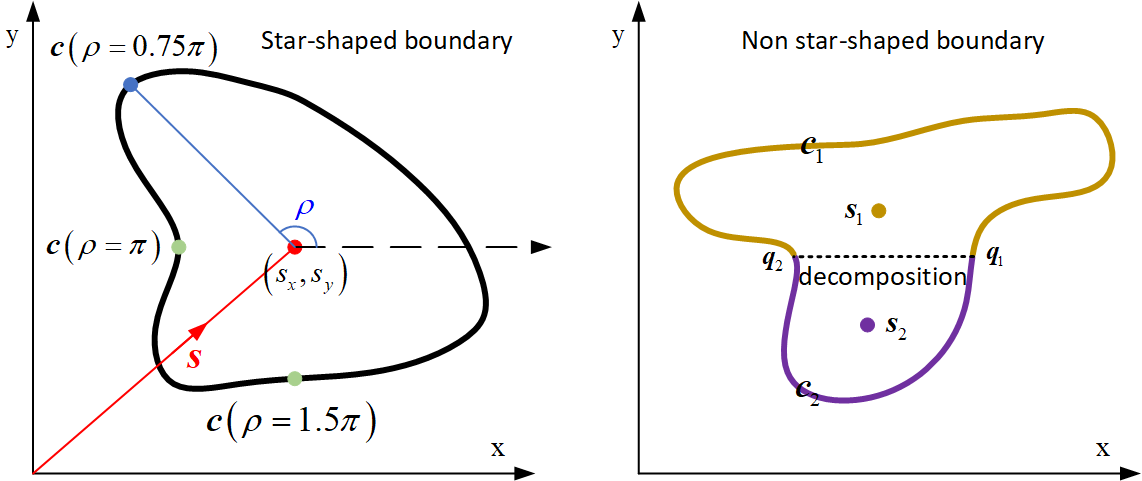}
    \caption{Boundary representation.}
    \label{boundary}
\end{figure}

We choose a polar angle as the parameter $\rho$ in \eqref{contour}, which is defined as
\begin{equation}\label{angle}
    \rho=\arctan\left(\frac{y-s_y}{x-s_x}\right)
\end{equation}
where $\bm{s}=[s_x,s_y]^\top$, as shown in Fig. \ref{boundary}. According to Definition \ref{def1}, $\bm{s}$ can be selected as the reference point of the boundary set. In this manner, the polar-angle-based parametric equation \eqref{contour} with \eqref{angle} can characterize all star-shaped boundaries, since there exists an injective function that maps all boundary points to different polar angles.
As for the non-star-shaped boundary, we may first decompose it into several star-shaped sets, as shown in the right subfigure of Fig. \ref{boundary}. Then, the boundary can be parameterized separately on these segments. 
Note that in this case, the parameter domain is no longer $\rho\in[0,2\pi)$, and it is determined by reference points and cut-off points between different segments. For example, as in the right subfigure of Fig. \ref{boundary}, two segments $\bm{c}_1(\rho)$ with $\rho\in[\arctan(\frac{s_{1y}-q_{1y}}{s_{1x}-q_{1x}}),\arctan(\frac{s_{1y}-q_{2y}}{s_{1x}-q_{2x}}))$, and $\bm{c}_2(\rho)$ with $\rho\in[\arctan(\frac{s_{2y}-q_{2y}}{s_{2x}-q_{2x}}),\arctan(\frac{s_{2y}-q_{1y}}{s_{2x}-q_{1x}}))$ need to be fitted, where $\bm{q}_1$, $\bm{q}_2$ are the cut-off points between the two segments.

\vspace{-0.5em}
\textit{
\begin{remark} 
    The notion of star-shaped set is a generalization of convex set. In other words, our proposed representation method \eqref{contour} with \eqref{angle} can characterize any convex boundary. In addition, it can also be applied to non-star-shaped boundaries by decomposition and piecewise representation.
\end{remark}}
\vspace{-0.5em}
\textit{
\begin{remark}\label{r2}
    There are some other choices for the parameter $\rho$ in \eqref{contour}, such as the arc length \cite{romero2022model,zhang2023fourier}. 
However, it is non-trivial to measure the distance to an arc-length-based parametric curve, because this requires solving an optimization problem which is even non-convex, i.e., 
    \begin{equation}
        \text{dist}(\xi,\bm{c}(\rho))=\min_\rho \ \sqrt{(\xi_x-c_x(\rho))^2+(\xi_y-c_y(\rho))^2}
    \end{equation}
    where $\text{dist}(\xi,\bm{c}(\rho))$ is the Euclidean distance from position $\bm{\xi}=[\xi_x,\xi_y]^\top$ to the boundary. Therefore, we choose the polar angle \eqref{angle}, instead of arc length, to parameterize the boundary curve. The advantage is that we can use this polar angle to introduce a polar radius error, which helps to measure the `distance' easily, as will be shown later in Sec. \ref{sec3.2}.
\end{remark}}

To fit the boundary curve \eqref{contour}, we employ Fourier series to obtain an approximation $\tilde{\bm{c}}(\rho)=[\tilde{c}_x(\rho),\tilde{c}_y(\rho)]^\top$, detailed as
\begin{subequations}\label{fit}
    \begin{align}
    \tilde{c}_x(\rho)&=\sum_{h=1}^{H}a_h\cos{h\rho}+b_h\sin{h\rho}+e \\ 
    \tilde{c}_y(\rho)&=\sum_{h=1}^{H}c_h\cos{h\rho}+d_h\sin{h\rho}+f
    \end{align}
\end{subequations}
where $H>0$ is the number of harmonic taking into account, $a_h$, $b_h$, $c_h$, $d_h$ are coefficients corresponding to the $h$-th harmonic, and $e$, $f$ are truncated errors. We group the fitting parameters into a vector, that is,
\begin{equation}
    \bm{\zeta}=[\bm{\eta}_1^\top,\bm{\eta}_2^\top,\cdots,\bm{\eta}_H^\top,e,f]^\top\in\mathbb{R}^{4H+2}
\end{equation}
with $\bm{\eta}_h=[a_h,b_h,c_h,d_h]^\top$. Then we can rewrite \eqref{fit} into a compact matrix form as
\begin{equation}\label{approximation}
    \tilde{\bm{c}}(\rho)=\bm{G}(\rho)\bm{\zeta}
\end{equation}
where $\bm{G}(\rho)=[\bm{g}_1(\rho), \bm{g}_2(\rho),\cdots,\bm{g}_H(\rho),\bm{I}_2]\in\mathbb{R}^{2\times(4H+2)}$ is a regression-like matrix that contains the harmonic terms, with
\begin{equation}
    \bm{g}_h(\rho)=\left[
    \begin{array}{cccc}
     \cos(h\rho) & \sin(h\rho) & 0 & 0 \\
     0 & 0 & \cos(h\rho) & \sin(h\rho)
\end{array} \right]
\end{equation}

The least square method can be used to obtain the optimal fitting parameter $\bm{\zeta}$ for the approximated boundary. Suppose there are $N$ sampled boundary points, and we stack them into the following point sequence
\begin{equation}
    \bm{C}=[\bm{c}(\rho_1)^\top,\bm{c}(\rho_2)^\top,\cdots,\bm{c}(\rho_N)^\top]^\top\in\mathbb{R}^{2N}
\end{equation}
Based on \eqref{approximation}, it yields that
\begin{equation}\label{regression}
    \tilde{\bm{C}}=\bm{\Gamma}\bm{\zeta}
\end{equation}
with
\begin{equation}
   \tilde{\bm{C}}=\left[ \begin{array}{c}
       \tilde{\bm{c}}(\rho_1) \\
       \vdots \\
       \tilde{\bm{c}}(\rho_N)
   \end{array}\right],\quad
   \bm{\Gamma}=\left[\begin{array}{c}
       \bm{G}(\rho_1) \\
       \vdots \\
       \bm{G}(\rho_N)
   \end{array}\right]
\end{equation}
Therefore, the fitting parameter $\bm{\zeta}$ is computed as
\begin{equation}
    \bm{\zeta}=\left(\bm{\Gamma}^\top\bm{\Gamma}\right)^{-1} \bm{\Gamma}^\top\bm{C}
\end{equation}

There are some considerations for choosing a proper $H$ for boundary approximation. Firstly, for a boundary with $N$ sampled points, $2H+1<N$ should be satisfied to ensure existence of $(\bm{\Gamma}^\top\bm{\Gamma})^{-1}$. Secondly, larger $H$ will increase approximation accuracy, however, it may introduce high-harmonic oscillations. In addition, there may be unwanted noise when collecting sampled boundary points from sensors, so we need to select a moderate $H$ to make a trade-off between improving precision and avoiding the over-fitting problem.

\subsection{VF-Guided Reference Control} \label{sec3.2}

\textit{
\begin{lemma}\cite{kapitanyuk2018guiding}\label{lemma}
    If the desired path is described by an implicit function $\phi(x,y)=0$, then a valid 2-D vector field is
    \begin{equation}\label{vf}
        \bm{\xi}(x,y)=\bm{\tau}(x,y)-k\delta(\phi(x,y))\bm{n}(x,y)
    \end{equation}
    where $\bm{\tau}$ is the tangent vector to the path, $\bm{n}$ is the normal vector, and $\delta(\phi(x,y))$ is considered as a ``signed'' distance to the path, with $\delta(\cdot)$ being a strictly increasing function.
    It is concluded that the integral curves of this field lead either to the desired path or to the critical set (where $\bm{\tau}(x,y)=\bm{0}$).
\end{lemma}}
\vspace{0.5em}

The first term in \eqref{vf} is tangential to the desired path, which can drive the robot to traverse the curve. The second term is perpendicular to the first term and related to the tracking error, and thus it steers the robot to converge to the desired path. 

\begin{figure}[t]
    \centering
    \includegraphics[width=8.5cm]{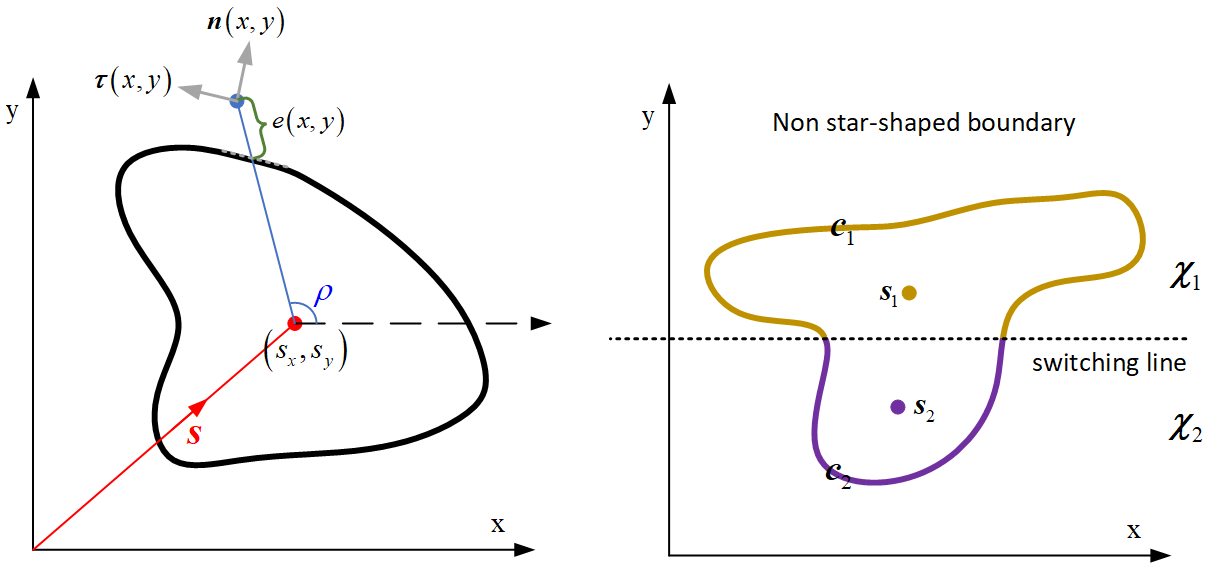}
    \caption{Left: Tangent vector and tracking error.  Right: Switching mechanism for non star-shaped boundary, and two VFs work in different regions.}
    \label{vector}
\end{figure}

We are now in a position to derive our VF for the approximated boundary \eqref{fit}. According to Lemma \ref{lemma}, the key point is: 1) to calculate the tangent direction; and 2) to find a `distance' to the boundary as a tracking error. For the former, it is obvious that the tangent vector $\bm{\tau}$ can be evaluated as
\begin{equation}
\begin{aligned}
    \bm{\tau}(x,y)=&\left[\frac{d\tilde{c}_x(\rho)}{d\rho},\frac{d\tilde{c}_y(\rho)}{d\rho}\right]^\top  \\
    = &\left[\sum_{h=1}^{H}-h\cdot a_h\sin{h\rho}+h\cdot b_h\cos{h\rho}, \right.\\
    &\left. \sum_{h=1}^{H}-h\cdot c_h\sin{h\rho}+h\cdot d_h\cos{h\rho}\right]^\top
\end{aligned}
\end{equation}
Then the normal vector $\bm{n}(x,y)$ is
\begin{equation}
     \bm{n}(x,y)=\bm{E}\bm{\tau}(x,y),\ \bm{E}=\left[\begin{array}{cc}
       0  & 1 \\
       -1  & 0
    \end{array}\right]
\end{equation}
The parametric equation \eqref{fit} can also be described into an implicit form as $(x-\tilde{c}_x(\rho))^2+(y-\tilde{c}_y(\rho))^2=0$, with $\rho$ computed in \eqref{angle} based on $x$ and $y$. 
Then we introduce a `signed' distance to the boundary, which is defined as
\begin{equation}\label{error}
    e(x,y)=\sqrt{(x-\tilde{c}_x(\rho))^2+(y-\tilde{c}_y(\rho))^2}
\end{equation}
which is actually the polar radius error from position $(x,y)$ to the approximated boundary, as shown in Fig. \ref{vector}.

The reference control for boundary encircling is then designed as
\begin{equation}\label{ref}
    \bm{u}_r=v_d\cdot\frac{\bm{\chi}(x,y)}{\|\bm{\chi}(x,y)\|}
\end{equation}
where $v_d>0$ is the desired encircling speed, and $\bm{\chi}(x,y)$ is the VF derived as
\begin{equation}
    \bm{\chi}(x,y)=\bm{\tau}(x,y)-ke(x,y)\bm{n}(x,y)
\end{equation}
with $k>0$ being a tunable parameter. 

Problem \ref{p1} stated in Sec. \ref{sec2.3} can be solved by using reference controller \eqref{ref}. With this control law, the robot is governed by $\dot{\bm{x}}=\bm{u}_r$. According to Lemma \ref{lemma}, the robot will reach and then encircle the boundary (without considering the critical points). It is also clear that the robot will move with a desired constant speed $v_d$, since $\|\bm{u}_r\|=v_d$. In addition, once the robot arrives at the boundary, its orientation will be steered along the tangent direction, as $\bm{\chi}=\bm{\tau}$ (which is the tangent vector of the boundary) after arrival. 

Furthermore, for non-star-shaped boundaries, we need to switch between different VFs to achieve encirclement. As shown in the right part of Fig. \ref{vector}, two VFs $\bm{\chi}_1$ and $\bm{\chi}_2$ should be designed and work in different regions separated by the switching line.

\vspace{-0.5em}
\textit{
\begin{remark}
    One of the drawbacks of the VF method is the existence of critical points where the robot may be trapped. 
    We can obtain these points numerically by solving $\bm{\tau}(x,y)=\bm{0}$. However, such points rarely exist for irregular boundaries as two equations $d\bar{c}_x(\rho)/d\rho=0$ and $d\bar{c}_y(\rho)/d\rho=0$ have to be satisfied. This problem can be alleviated by exerting a small additive velocity to help the robot escape from critical points.
\end{remark}}

\vspace{-0.5em}
\textit{
\begin{remark}
    A stand-off distance $e_d$ can be set for boundary encircling by changing \eqref{error} into $e=\sqrt{(x-\tilde{c}_x(\rho))^2+(y-\tilde{c}_y(\rho))^2}-e_d$. In this way, the robot keeps a constant polar radius distance to the boundary during encirclement, which can enlarge its patrolling area.
\end{remark}}

\subsection{Controller Synthesis Subject to Constraints}
Obstacle avoidance and input saturation constraints are commonly seen in practical engineering and the reference control $\eqref{ref}$ cannot be used directly in these cases. We need to modify it to accommodate these constraints so that Problem \ref{p2} can be addressed.
CBF is an effective method to deal with system constraints. It constructs a forward-invariant set to ensure that the states evolve within a constrained safety set. Input constraints can also be addressed by incorporating them into a QP problem. 

\vspace{-0.5em}
\textit{
\begin{definition}
    (Control barrier function (CBF))\cite{li2023survey}. Consider a closed set $\mathcal{C}=\{\bm{x}\in \mathbb{R}^n|h(\bm{x})\geq0\}$ with boundary $\partial\mathcal{C}=\{\bm{x}\in \mathbb{R}^n|h(\bm{x})=0\}$ and interior $Int(\mathcal{C})=\{\bm{x}\in \mathbb{R}^n|h(\bm{x})>0\}$. For an affine-control system $\dot{\bm{x}}=\bm{f}(\bm{x})+\bm{g}(\bm{x})\bm{u}$ with state $\bm{x}\in\mathbb{R}^n$ and input $\bm{u}\in\mathbb{R}^m$, the function $h(\bm{x})$ is a valid CBF if there exists an extended class $\mathcal{K}$ function $\alpha(\cdot)$ such that
    \begin{equation}
    \sup_{\bm{u}\in\mathcal{U}}\left[\frac{\partial h}{\partial\bm{x}}\bm{f}(\bm{x})+\frac{\partial h}{\partial\bm{x}}\bm{g}(\bm{x})+\alpha(h(\bm{x}))\right]\geq0
    \end{equation}
\end{definition}}
\vspace{0.5em}

The above equation guarantees the forward-invariance of set $\mathcal{C}$ along system trajectory, because it yields that $\dot{h}(\bm{x})\geq-\alpha(h(\bm{x}))$. Therefore, for any $\bm{x}(t_0)\in\mathcal{C}$, we have $\bm{x}(t)\in\mathcal{C}$. In this way, the states will remain within the specific set $\mathcal{C}$.
Note that a special case for function $\alpha(h(\bm{x}))$ is the linear form $\alpha(h(\bm{x}))=\alpha h(\bm{x})$ with $\alpha>0$.

Based on the above CBF theory, the following function is constructed for obstacle avoidance constraint \eqref{state}
\begin{equation}\label{h}
    h(\bm{x})=\|\bm{x}-\bm{x}_o^i\|^2-(r_o^i)^2,\ i=1,2,\cdots,n
\end{equation}
and it is required that
\begin{equation}
    \dot{h}(\bm{x})=2(\bm{x}-\bm{x}_o^i)^\top\bm{u}\geq-\alpha h(\bm{x})
\end{equation}
where $\alpha>0$ is a tunable parameter. In addition, input constraint \eqref{input} can be directly encoded.
Then we formulate the following QP problem to modify the reference control \eqref{ref} in a minimally invasive way 
\begin{equation}\label{control}
\begin{aligned}
    \bm{u}=&\mathop{\arg\min}\limits_{\bm{u}}\ \|\bm{u}-\bm{u}_r\|^2 \\
    \text{s.t.}\enspace &2(\bm{x}-\bm{x}_o^i)^\top\bm{u}\geq-\alpha h(\bm{x}) \\
    &-\bm{b}\leq\bm{AR}^{-1}\bm{u}\leq\bm{b}
\end{aligned} 
\end{equation}

By solving this QP problem for control input $\bm{u}$, Problem \ref{p2} can be solved. Intuitively, the QP formulation serves as a safety filter that minimally corrects the infeasible reference control $\bm{u}_r$. The solution $\bm{u}$ is actually the nearest control law around $\bm{u}_r$ that can ensure the satisfaction of hard constraints described by \eqref{state} and \eqref{input}.

\section{Results}

In this section, we validate our control methodology with both numerical simulations and hardware experiments. 
A video demonstrating the performance of the proposed method is available at \url{https://vimeo.com/1043607338}.

\subsection{Numerical Simulations}
\begin{figure*}[t]
	\centering
	\subfigure[Case 1.]{
		\includegraphics[height=5.1cm]{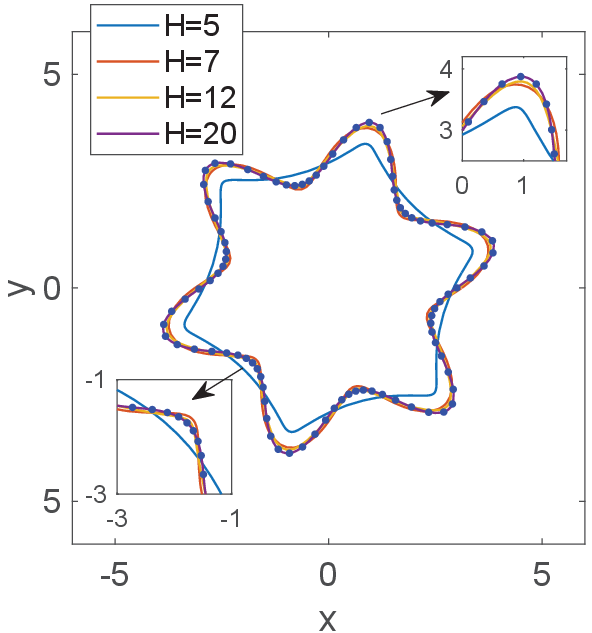}
		\label{case1fitting}}
	\subfigure[Case 2.]{
		\includegraphics[height=4.9cm]{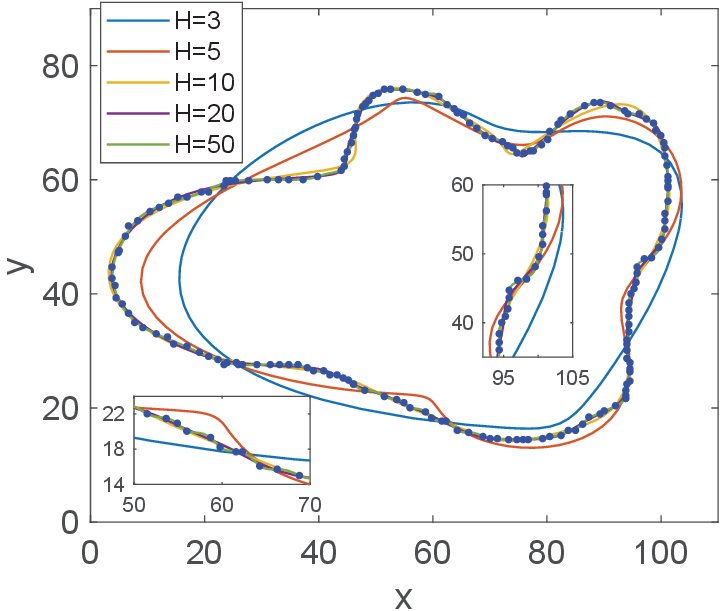}
		\label{case2fitting}}
	\subfigure[Case 3.]{
		\includegraphics[height=4.9cm]{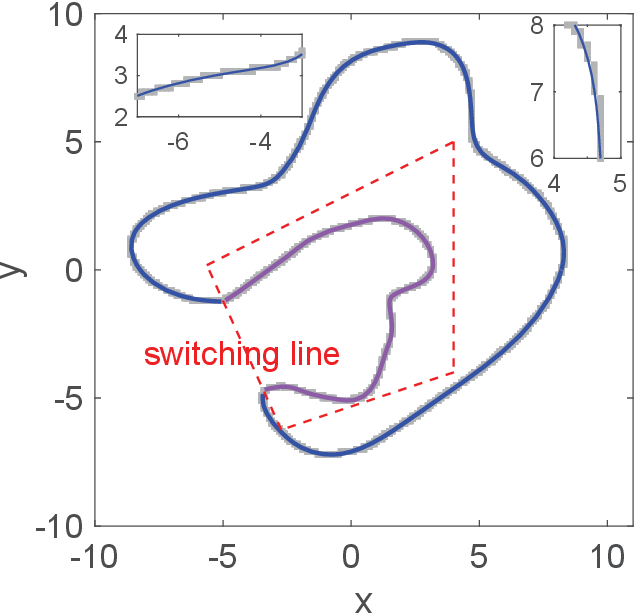}
		\label{case3fitting}} 
	\caption{Boundaries and approximated curves. In cases 1 and 2, the boundary is generated using discrete sampling points. In case 3, the boundary is extracted from an image using the edge detection algorithm.}
	\label{sim}
\end{figure*}

The mobile robot configuration is set as $l=0.01$, $d=0.3$, and $\alpha=1$ is set for obstacle avoidance. Other control parameters are detailed in Tab. \ref{tab1}. 
Three cases are simulated: 1) regular star-shaped boundary, 2) irregular boundary with sampling noise, and 3) non-star-shaped boundary. Sampled points and approximated curves are presented in Fig. \ref{sim}.

\begin{table}[t]
	\caption{Control parameters for simulation.}
	\label{tab1}
	\centering
	\begin{threeparttable}
		\begin{tabular*}{250pt}
			{@{\extracolsep\fill}lccccc@{\extracolsep\fill}}%
			\toprule
			Case  & $k$ & $v_d$ & $v_m$ & $\bm{x}_o$ & $r_o$\\
			\midrule
			1   & 1   & 0.2 & 0.3 & [2.6 1.4], [0 -3] & 0.5, 0.3 \\
			2   & 0.1 & 2   & 3   & [15 10], [15 30], [75 15], [95 75] & 5, 6, 8, 3\\ 
			3   & 1   & 0.5 & 1   & [-8 0], [3.5 0] & 1, 1 \\
			\bottomrule
		\end{tabular*}
	\end{threeparttable}
\end{table}

In case 1, the boundary points are sampled from the closed curve $r=2+2^{\sin{6\theta}}$, where $r$ and $\theta$ are polar radius and polar angle, respectively.
In case 2, the boundary points are generated with sampling noise and are plotted as blue dots. 
In addition, a non-star-shaped boundary is taken into consideration for case 3, and we need to approximate it separately on different segments. It is shown in Fig. \ref{case3fitting} that, blue and purple curves are the approximation for two boundary segments, where $H=15$ is used for the approximation. Gray dots are the extracted boundary, and they can be seen more clearly in the zoom-in figure. The switching line is plotted in red and it divides the space into two regions, each including a boundary segment. 

\begin{figure}[t]
	\centering
	\subfigure[Under $\bm{u}_r$ and $\bm{u}$.]{
		\includegraphics[height=4.9cm]{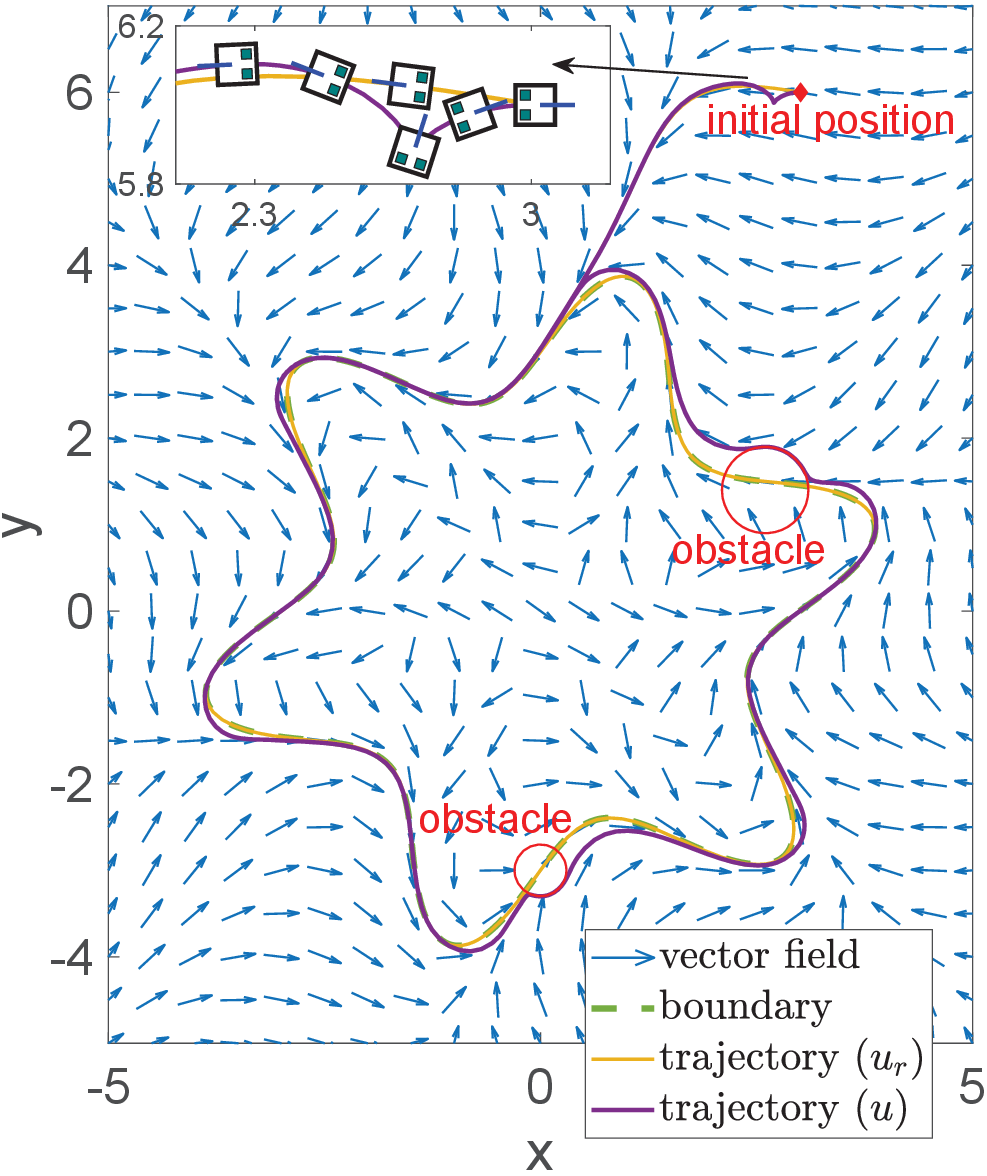}
		\label{case1traj}}
        \hspace{-1em}
	\subfigure[From different positions.]{
		\includegraphics[height=4.9cm]{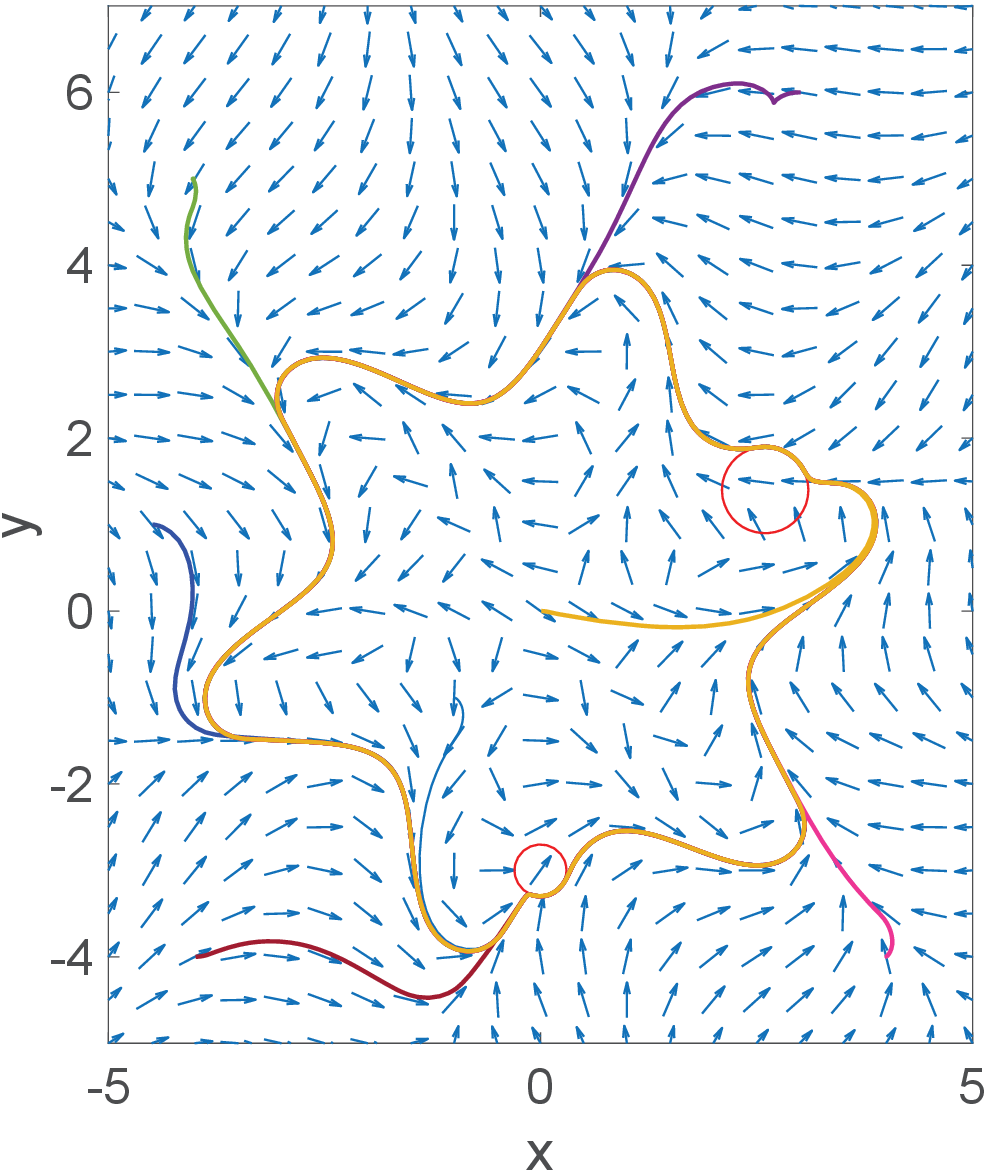}
		\label{case1initial}}
	\caption{Case 1: robot trajectories and the vector field.}
\end{figure}

\textit{Case 1: regular star-shaped boundary}. 
The trajectory starting from $p_x=3$, $p_y=6$ and $\theta=0$ is depicted in Fig. \ref{case1traj}. As can be seen, with the guidance of the designed vector field (blue arrows), the robot is steered to reach and then track the boundary. 
Specifically, under the reference control law $\bm{u}_r$, the robot can track the boundary accurately, as shown with the yellow curve. The purple curve in Fig. \ref{case1traj} is the trajectory using synthesized controller $\bm{u}$, which differs from the yellow curve at initial moments, during six sharp turns, and when bypassing obstacles. This is because the initially large tracking error $e$ results in a large reference control, which cannot be generated using limited wheel speed. As can be seen from the zoomed-in part, the robot turns around quickly under $\bm{u}_r$ at the initial moment, which actually requires high angular velocity that exceeds the robot's actuation ability. In contrast, the robot moves backward first and changes its heading gradually under the control input $\bm{u}$. Similarly, the robot motion trajectory at six sharp turns also shows a deviation of the purple curve from the yellow one, which is also caused by limited wheel speed $v_m$.
Besides, by modifying reference control $\bm{u}_r$ into control law $\bm{u}$, the robot can bypass two obstacles on the boundary successfully and continue encircling after obstacle avoidance. 

\begin{figure}[t]
	\centering
	\subfigure[Robot states.]{
		\includegraphics[width=4cm]{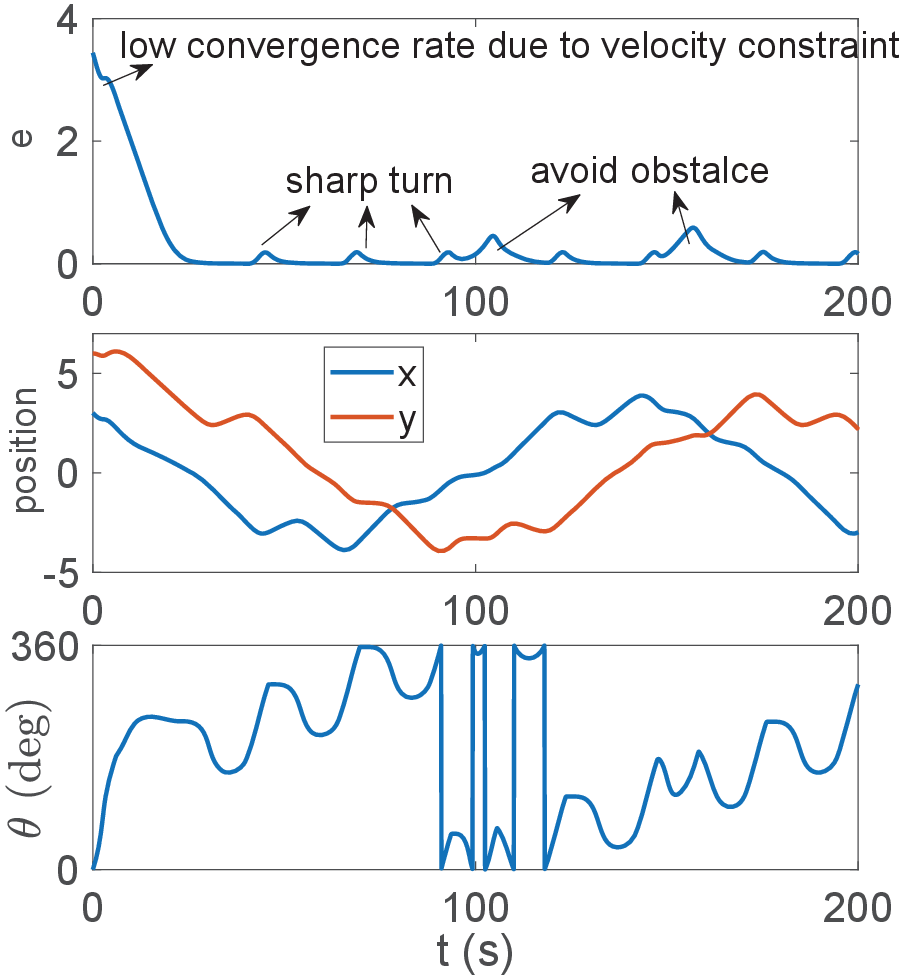}
		\label{case1state}}
	\subfigure[Robot velocities.]{
		\includegraphics[width=4.2cm]{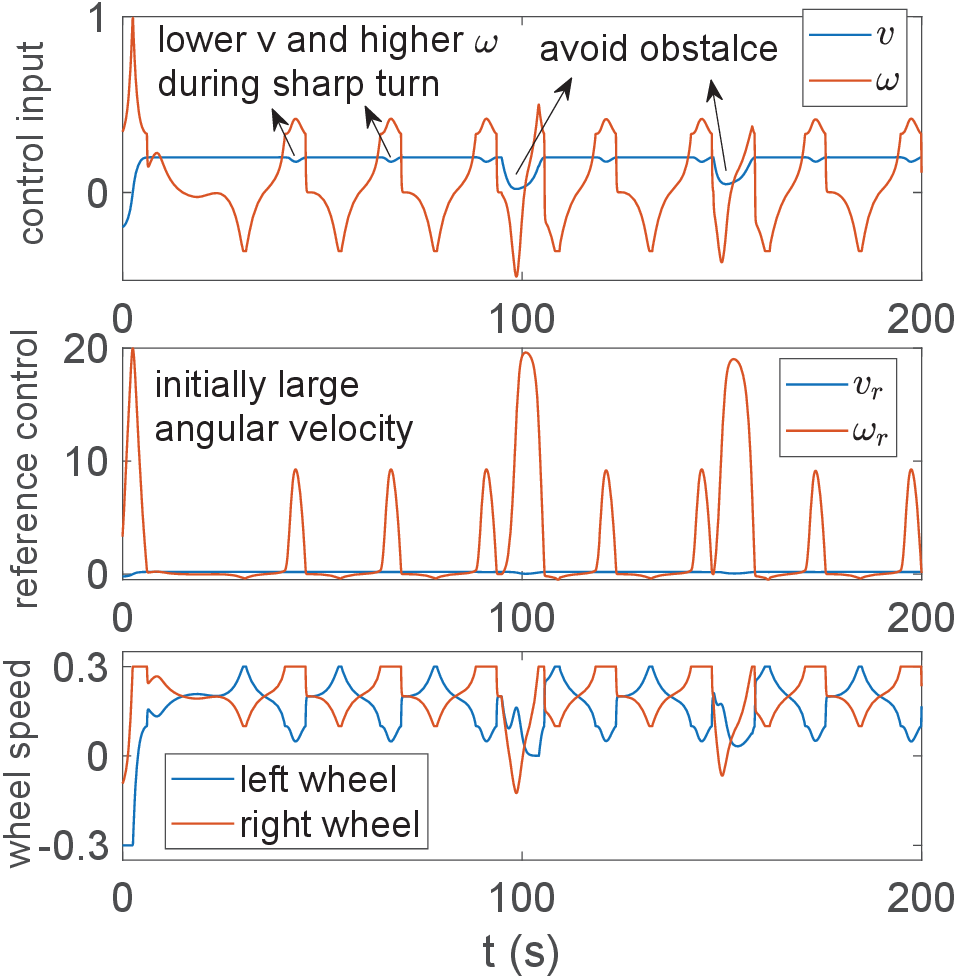}
		\label{case1input}}
	\caption{Case 1: time histories of system responses.}
\end{figure}

Fig. \ref{case1initial} further simulates more trajectories, and with our proposed control scheme $\bm{u}$, the robot achieves boundary encirclement from different initial positions.
In addition, the boundary tracking error $e$ is depicted in Fig. \ref{case1state}, which decreases to zero and shows fluctuations during sharp turn and obstacle-avoidance process. Control input $\bm{u}$ is shown in the upper of Fig. \ref{case1input}, wherein robot velocity $v$ keeps the constant value $v_d$ and lowers down during sharp turn and obstacle-avoidance process. Another notable thing is that the right wheel speed reaches maximum value $v_m$ during the six sharp turns, in order to change its heading quickly.


\begin{figure}[t]
	\centering
	\subfigure[With a stand-off distance.]{
		\includegraphics[height=3.9cm]{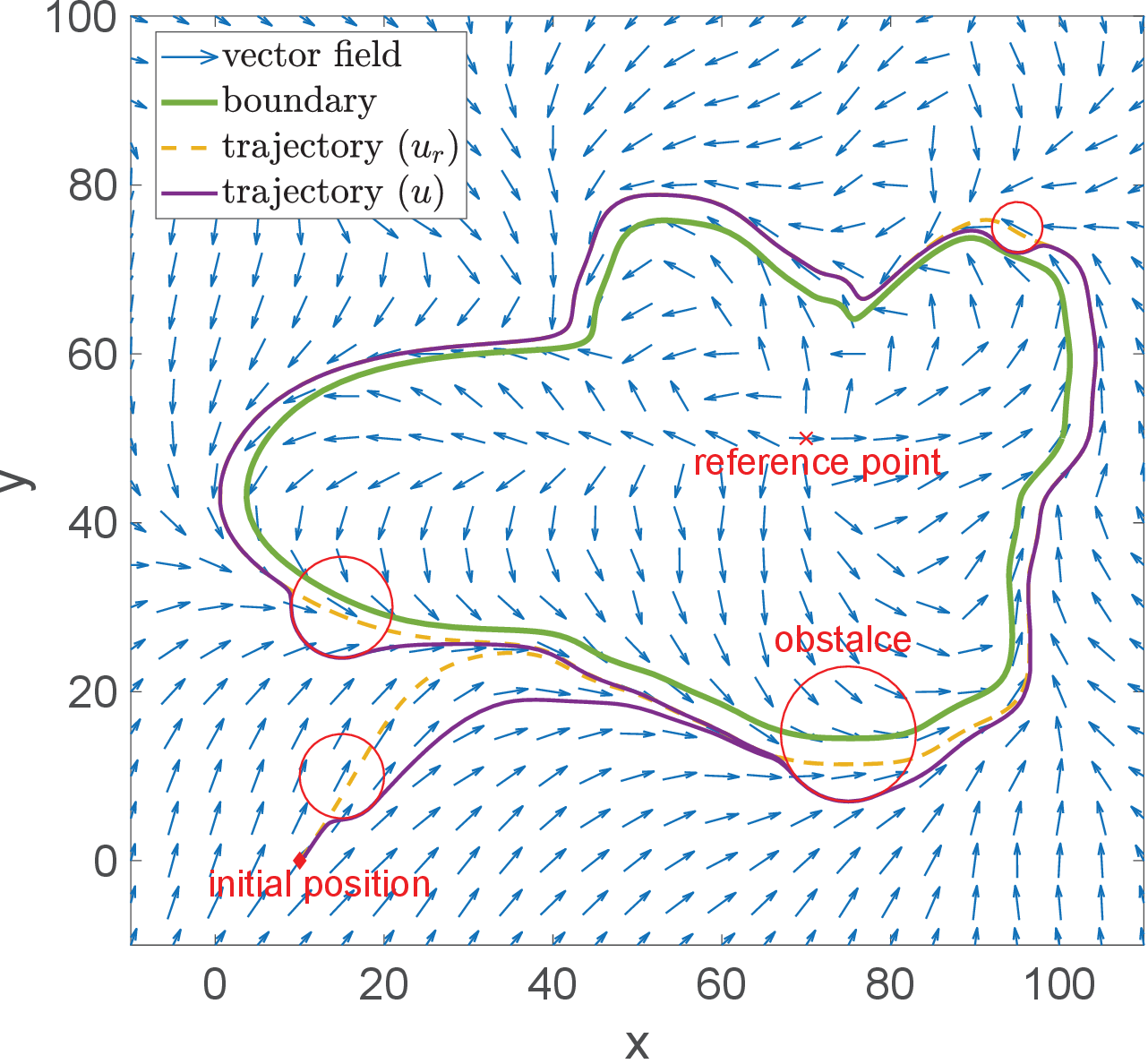}
		\label{case2traj}}
        \hspace{-1em}
	\subfigure[From different positions.]{
		\includegraphics[height=3.9cm]{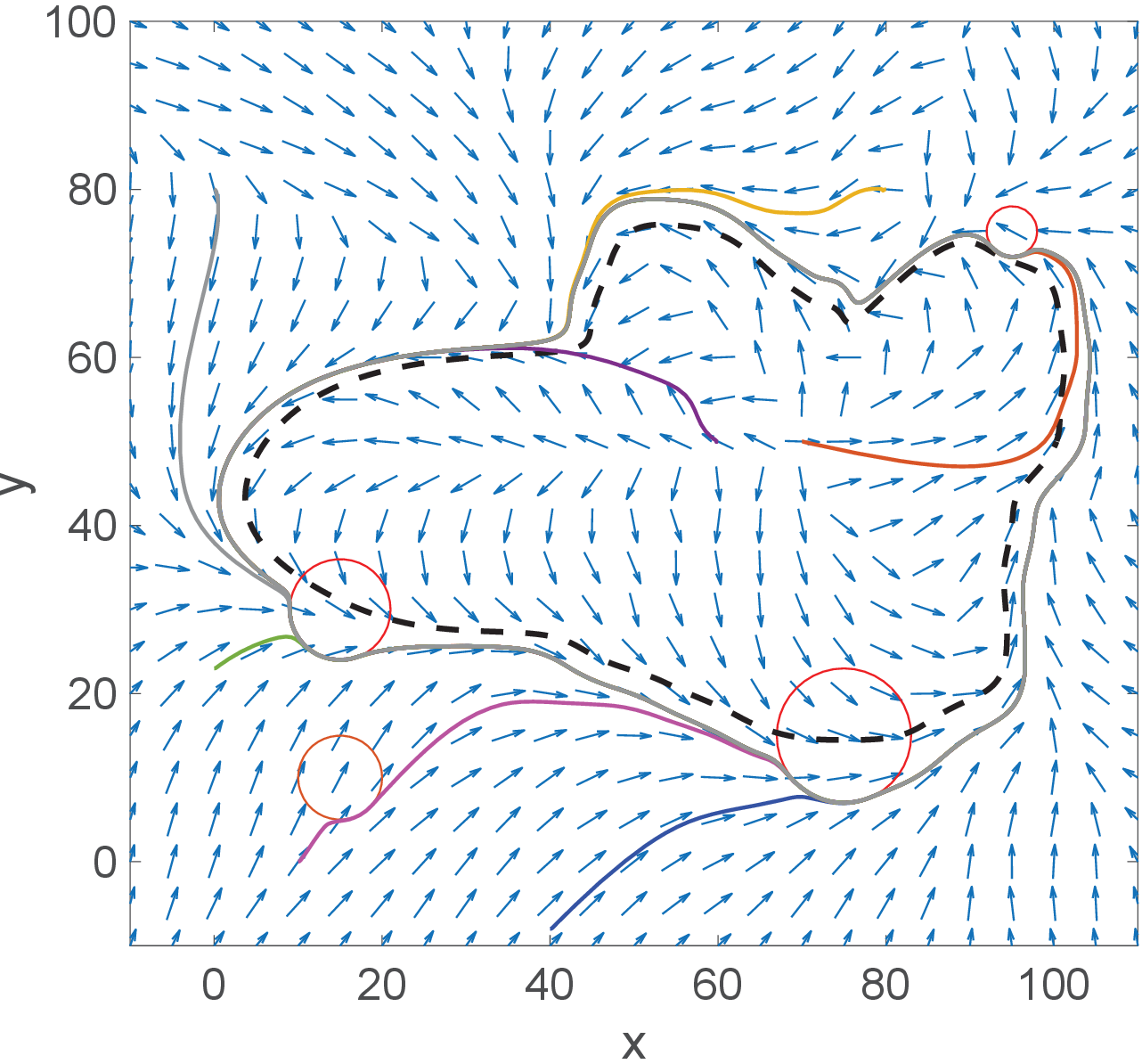}
		\label{case2initial}}
	\caption{Case 2: robot trajectories and the vector field.}
\end{figure}

\textit{Case 2: irregular boundary with sampling noise}. 
Fig. \ref{case2traj} shows the trajectories starting from $p_x=10$, $p_y=0$ and $\theta=0$ under control law \eqref{ref} and \eqref{control}, respectively. The reference point $\bm{s}$ of the boundary is shown in the cross mark, and four obstacles in Tab. \ref{tab1} are plotted with the dashed red circle. It is seen that the robot can encircle the boundary while avoiding obstacles using our proposed controller \eqref{control}. It is noted that we set a stand-off distance as $e_d=3$ in the simulation, and thus the robot keeps a constant polar radius distance to the boundary during its encirclement. This can expand the robot's patrolling area, which is useful in many applications.
Fig. \ref{case2initial} shows trajectories from different initial positions, wherein the dashed black line represents the fitted boundary. 

\begin{figure}[t]
	\centering
	\subfigure[Robot states.]{
		\includegraphics[width=4.3cm]{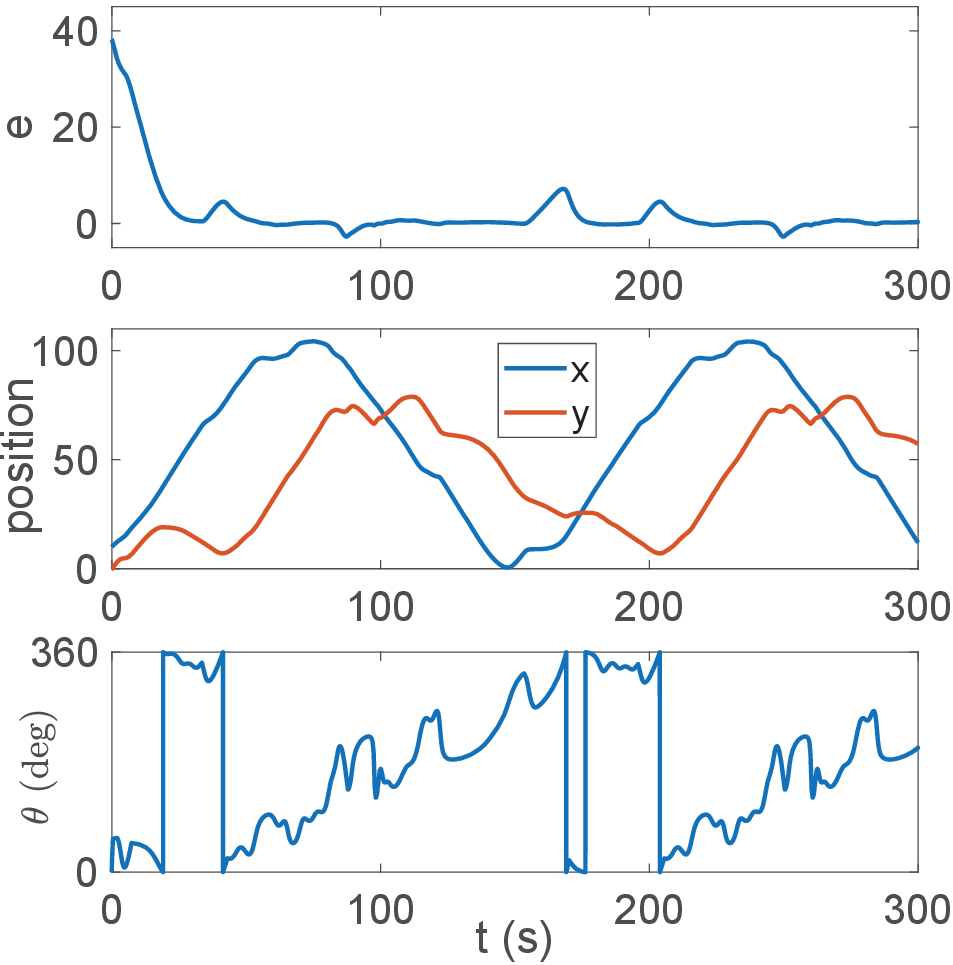}
		\label{case2state}}
	\hspace{-1em}
	\subfigure[Robot velocities.]{
		\includegraphics[width=4.1cm]{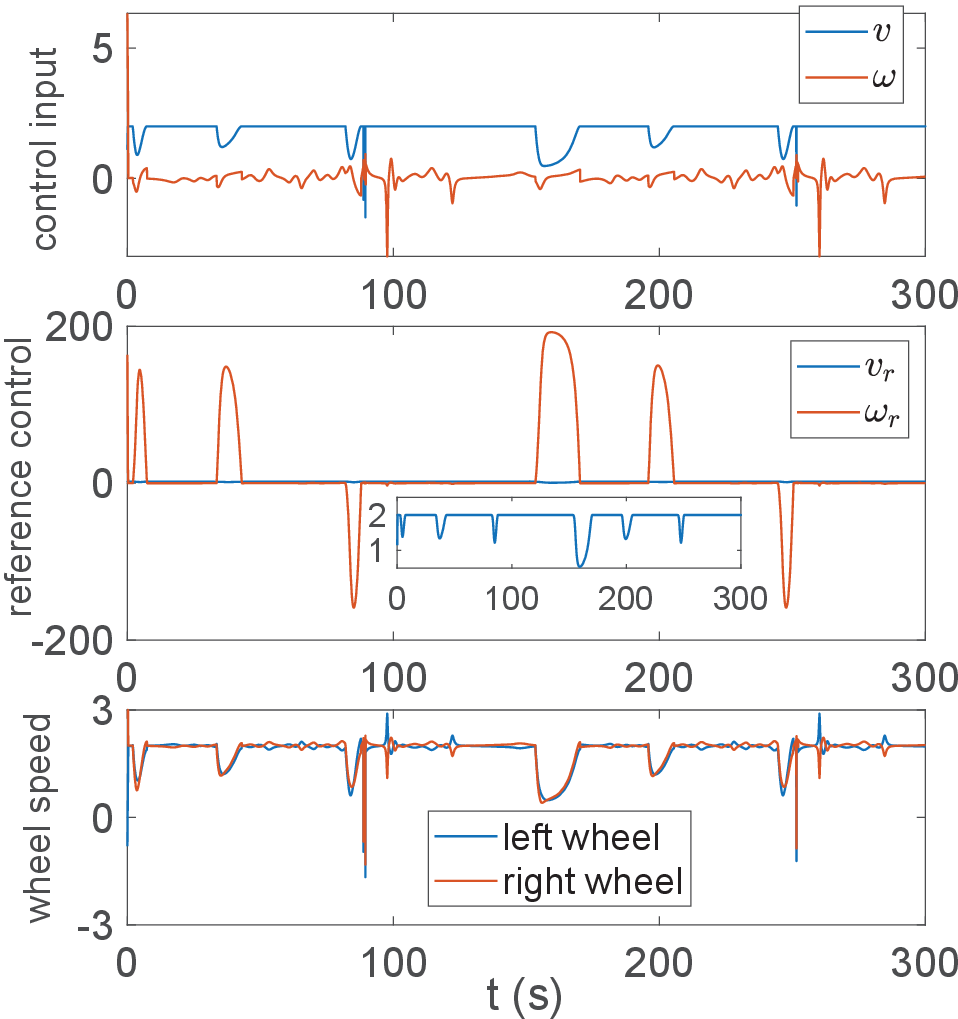}
		\label{case2input}}
	\caption{Case 2: time histories of system responses.}
\end{figure}

In addition, time histories are shown in Figs. \ref{case2state}-\ref{case2input}.
The polar radius error $e$ monotonically decreases to zero, and the fluctuations infer temporary deviations of the robot from the boundary, which is for obstacle avoidance. The robot states $x$, $y$, and $\theta$ are all periodic responses after it reaches the boundary, because the encirclement will go round and round. The control input $\bm{u}$ modifies reference control $\bm{u}_r$ using limited actuation. Therefore, although the large reference angular rate violates velocity constraints, the robot can still achieve the control objective using synthesized control input. It is shown that the robot moves at the desired constant speed and lowers its speed during obstacle avoidance to make a turn.


\begin{figure}[t]
	\centering
	\subfigure[Under $\bm{u}_r$ and $\bm{u}$.]{
		\includegraphics[height=4.1cm]{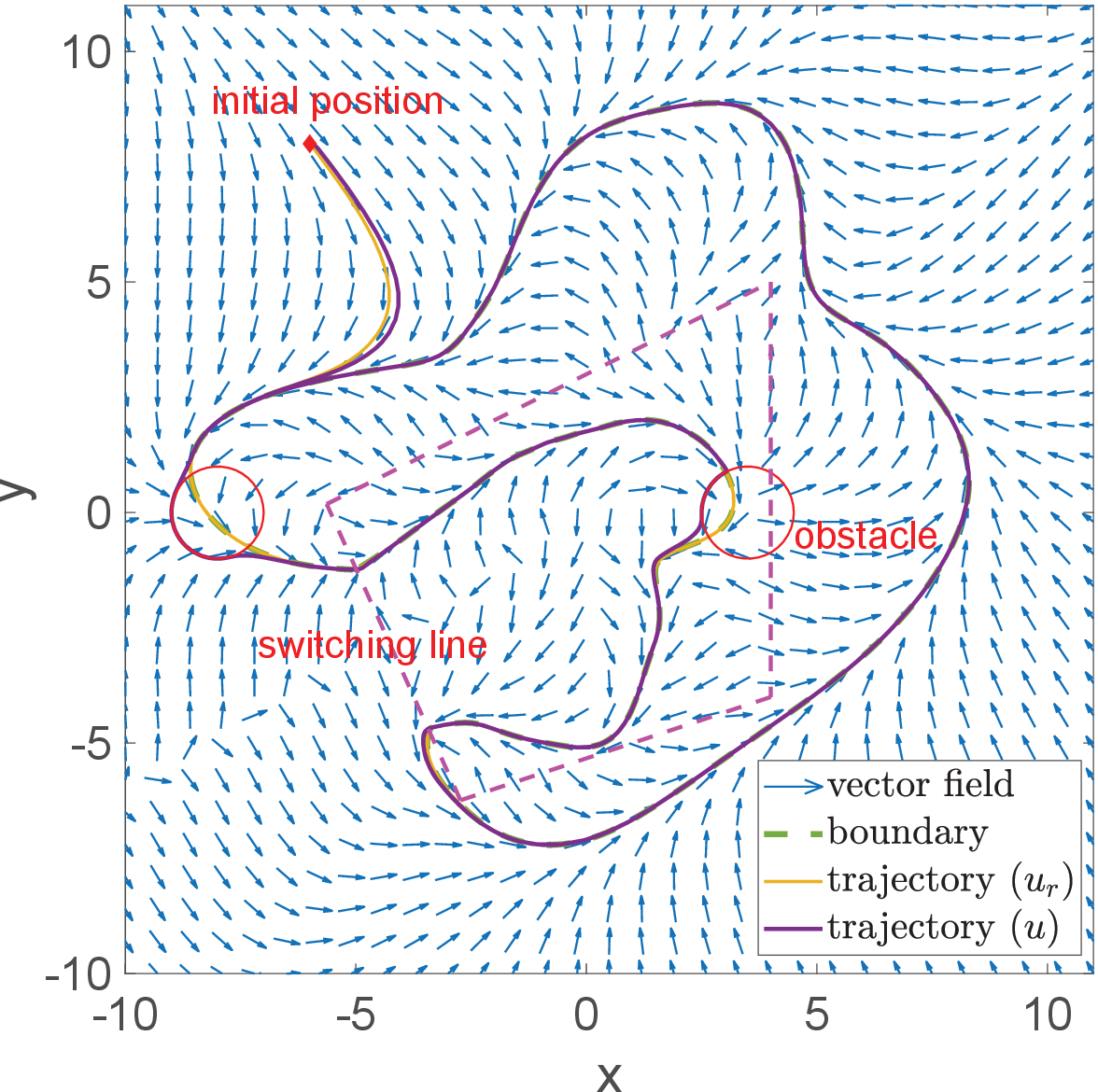}
		\label{case3traj}}
        \hspace{-1em}
	\subfigure[From different positions.]{
		\includegraphics[height=4.1cm]{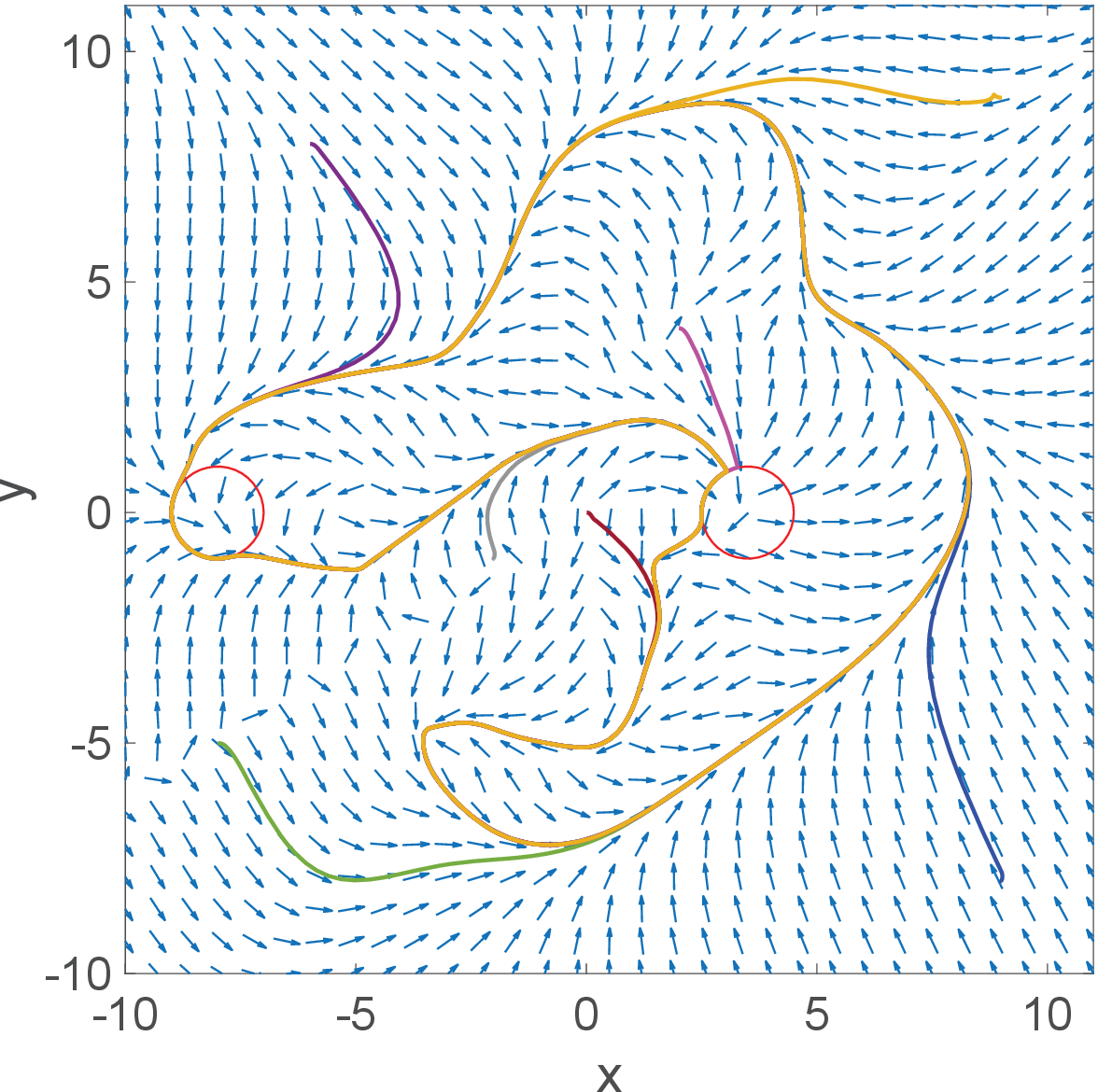}
		\label{case3initial}}
	\caption{Case 3: robot trajectories and the vector field.}
\end{figure}
\begin{figure}[t]
	\centering
	\subfigure[Robot states.]{
		\includegraphics[width=4.1cm]{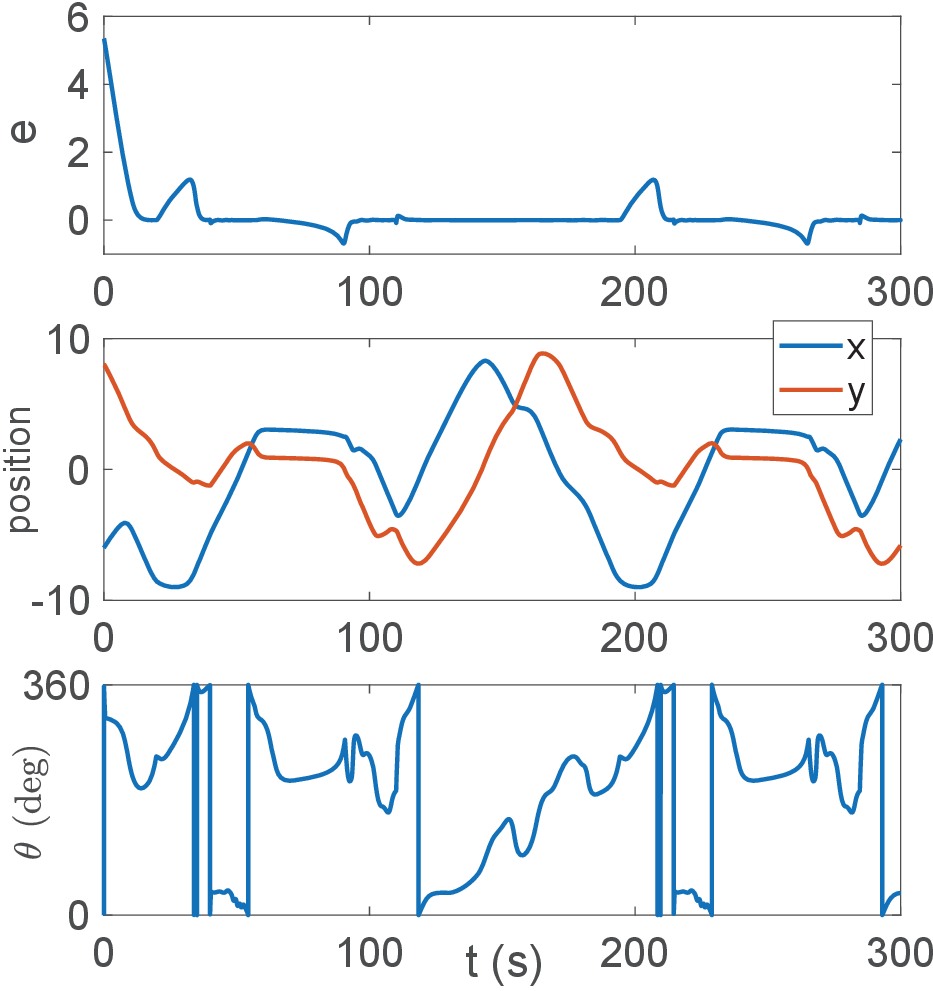}
		\label{case3state}}
	\subfigure[Robot velocities.]{
		\includegraphics[width=4.0cm]{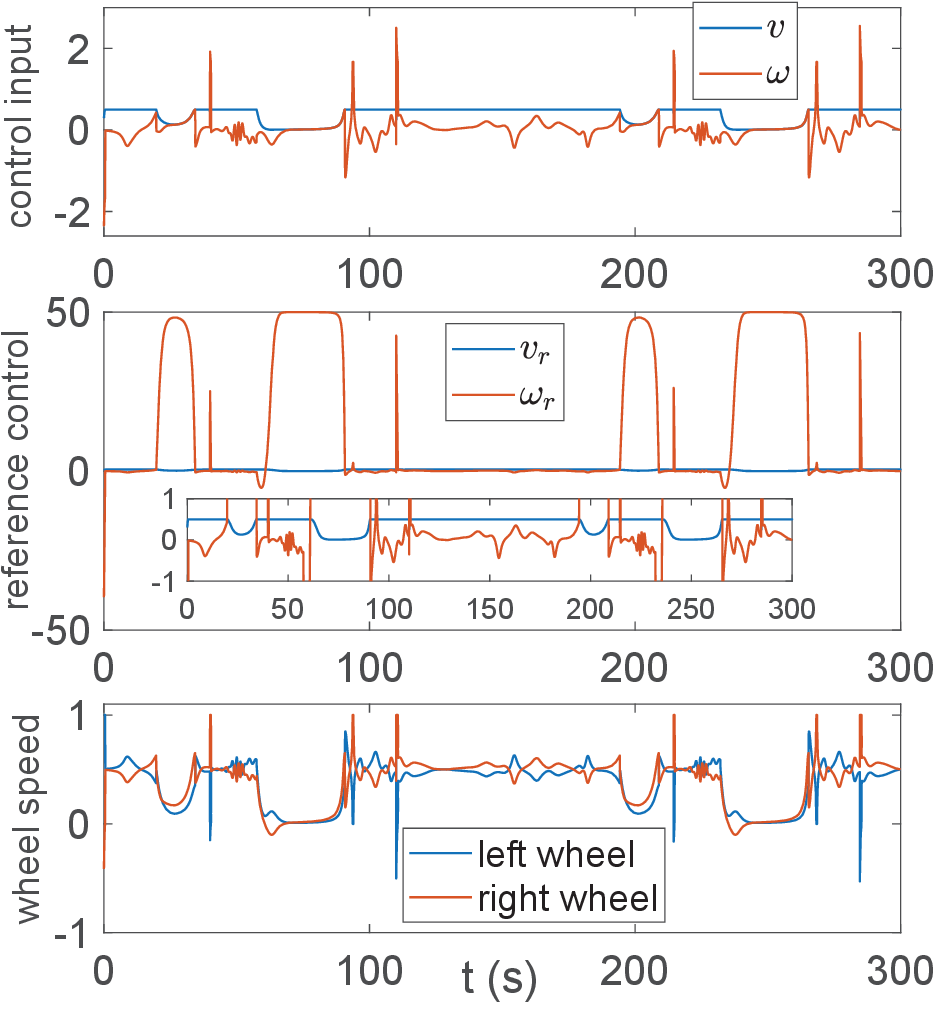}
		\label{case3input}}
	\caption{Case 3: time histories of system responses.}
\end{figure}

\textit{Case 3: non-star-shaped boundary}. 
The vector field for the approximated boundary is plotted in Fig. \ref{case3traj} with blue arrows. Note that the outer part (outside the switching line) is to be encircled anti-clockwise, while the inter part (inside the switching line) should be tracked clockwise, so that the trajectory will be continuous. 
The trajectory starting from $p_x=-6$, $p_y=8$ and $\theta=0$ is shown in Fig. \ref{case3traj}, and two trajectories under $\bm{u}_r$ and $\bm{u}$ are both simulated for comparison. Fig \ref{case3initial} plots trajectories from different initial positions, and Figs. \ref{case3state}-\ref{case3input} show the time histories of robot states and inputs, and relevant analysis is similar to cases 1 and 2.

\subsection{Experimental Results}
\begin{figure}[ht]
	\centering
	\includegraphics[width=8.5cm]{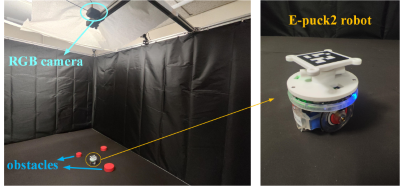}
	\caption{Experiment platform and the e-puck2 mobile robot.}
	\label{platform}
\end{figure}

We also conduct some experiments to verify the performance of our proposed control algorithm. The platform is shown in Fig. \ref{platform}, which is 2.8m$\times$1.4m in size. 
E-puck2 robot \cite{mondada2009puck} is used for the experiment, which is controlled using ROS by PC. Aruco marker is attached to the robot for positioning using the top-view camera. In addition, the red cylinders represent obstacles, and their position and radius can be obtained by the top-view RGB camera using OpenCV. Wi-Fi communication is built to send control commands and receive feedback data at a frequency of 10Hz between the PC and the robot. During the experiment, time delay caused by camera signal transmission, and unwanted disturbance from measurement noise are the two key factors decreasing the control performance. However, it is shown that our algorithm still achieves satisfactory results. 

\begin{figure}[t]
	\centering
	\subfigure[Experiment 1.]{
		\includegraphics[height=3.1cm]{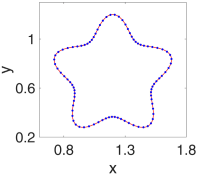}
		\label{ex1point}}
	\subfigure[Experiment 2.]{
		\includegraphics[height=3.1cm]{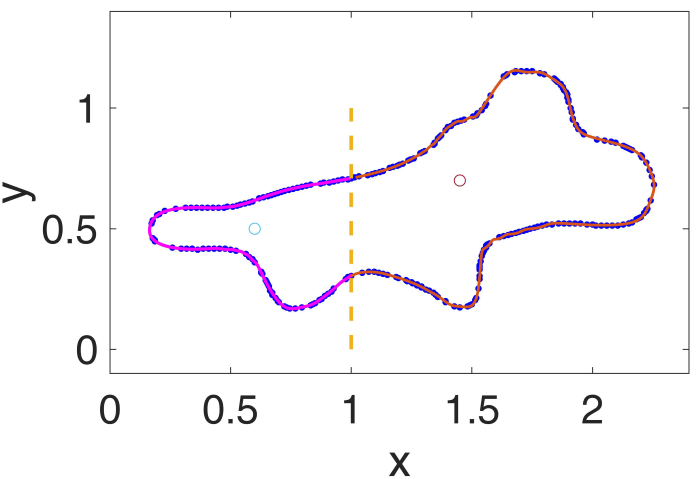}
		\label{ex2point}}
	\caption{Boundary points and approximated curves.}
	\label{point}
\end{figure}

Three experiments are conducted: in the first experiment, a star-shaped boundary is taken into consideration; non-star-shaped boundaries are considered for experiments 2 and 3.
The boundary points and approximated curves for experiments 1 and 2 are shown in Fig. \ref{point}, and we decompose the boundary of experiment 2 into two segments (separated by the dashed line) to perform curve fitting. 

\begin{figure}[ht]
	\centering
	\includegraphics[width=8.5cm]{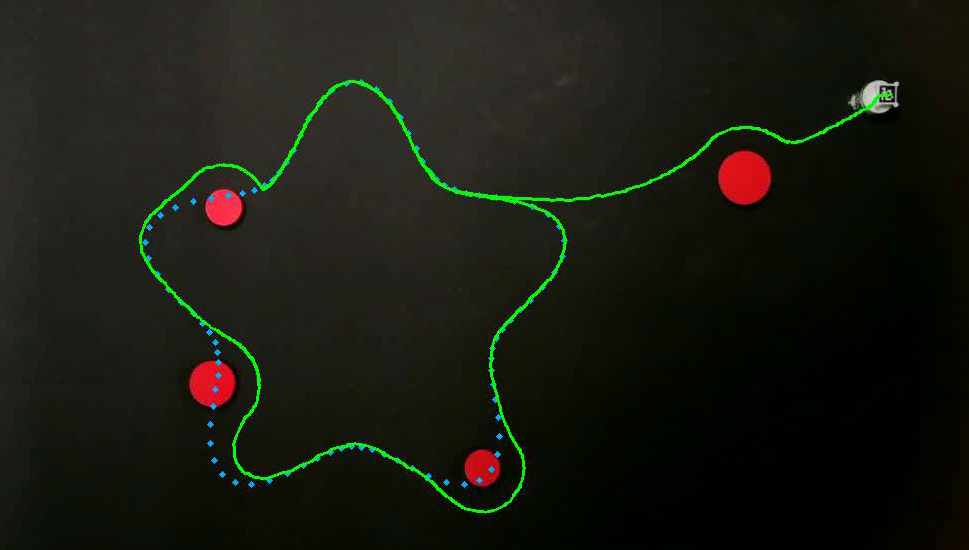}
	\caption{Robot trajectory in experiment 1.}
	\label{exp1traj}
\end{figure}

\begin{figure}[ht]
	\centering
		\includegraphics[width=8.5cm]{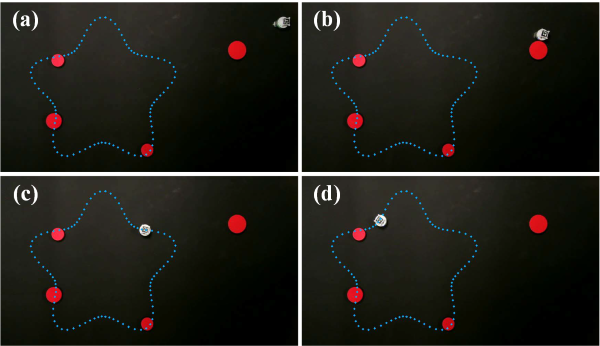}
        \caption{Four snapshots in experiment 1.}
	\label{exp1snap}
\end{figure}

\begin{figure}[ht]
	\centering
	\subfigure[Robot states.]{
		\includegraphics[width=3.9cm]{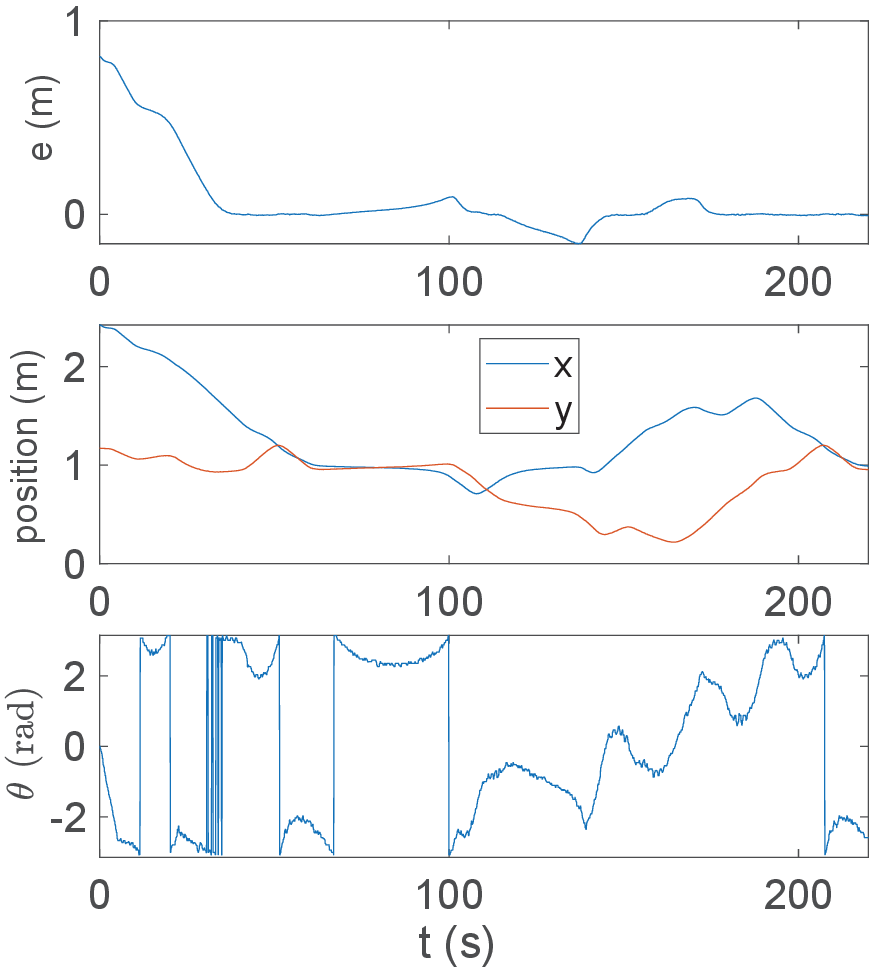}
		\label{case1position}}
	\subfigure[Robot velocities.]{
		\includegraphics[width=4.0cm]{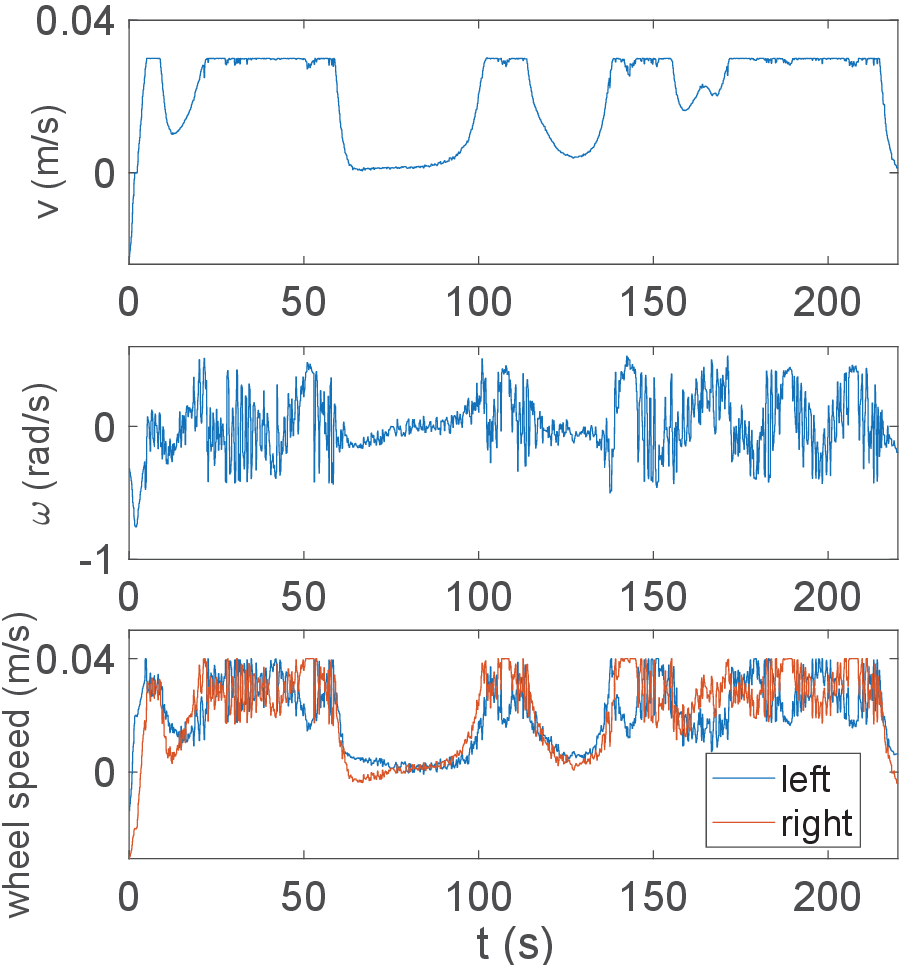}
		\label{case1velocity}}
	\caption{Time histories in experiment 1.}
	\label{exp1}
\end{figure}

In the first experiment, the robot starts from $[2.4,1.2]^\top$ m with an orientation of $0$ deg; one obstacle is placed on its way to the boundary, and three other obstacles are set on the boundary. Control parameters are taken as $k=10$, $v_d=0.03$ m/s, $v_m=0.04$ m/s, and the robot radius is $r_b=0.05$ m. In Fig. \ref{exp1traj}, the green line represents the motion trajectory of the robot center, and blue dots are sampled boundary points. It is shown that the robot can reach and then track the boundary and successfully avoids collision with obstacles throughout the whole process. Four snapshots during experiment 1 are presented in Fig. \ref{exp1snap}, and time histories are plotted in Fig. \ref{exp1}. 


\begin{figure}[ht]
	\centering
	\includegraphics[width=8.5cm]{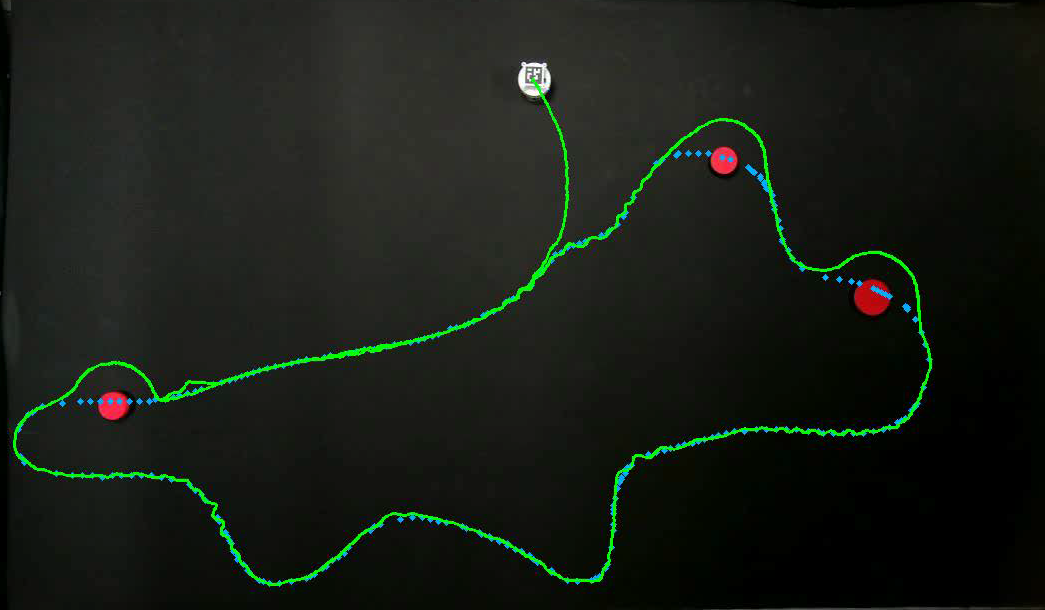}
	\caption{Robot trajectory in experiment 2.}
	\label{exp2traj}
\end{figure}
\begin{figure}[ht!]
	\centering
		\includegraphics[width=8.5cm]{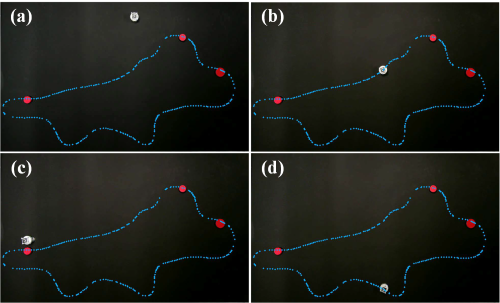}
	\caption{Four snapshots in experiment 2.}
	\label{exp2snap}
\end{figure}

\begin{figure}[t]
	\centering
	\subfigure[Robot states.]{
		\includegraphics[width=4.0cm]{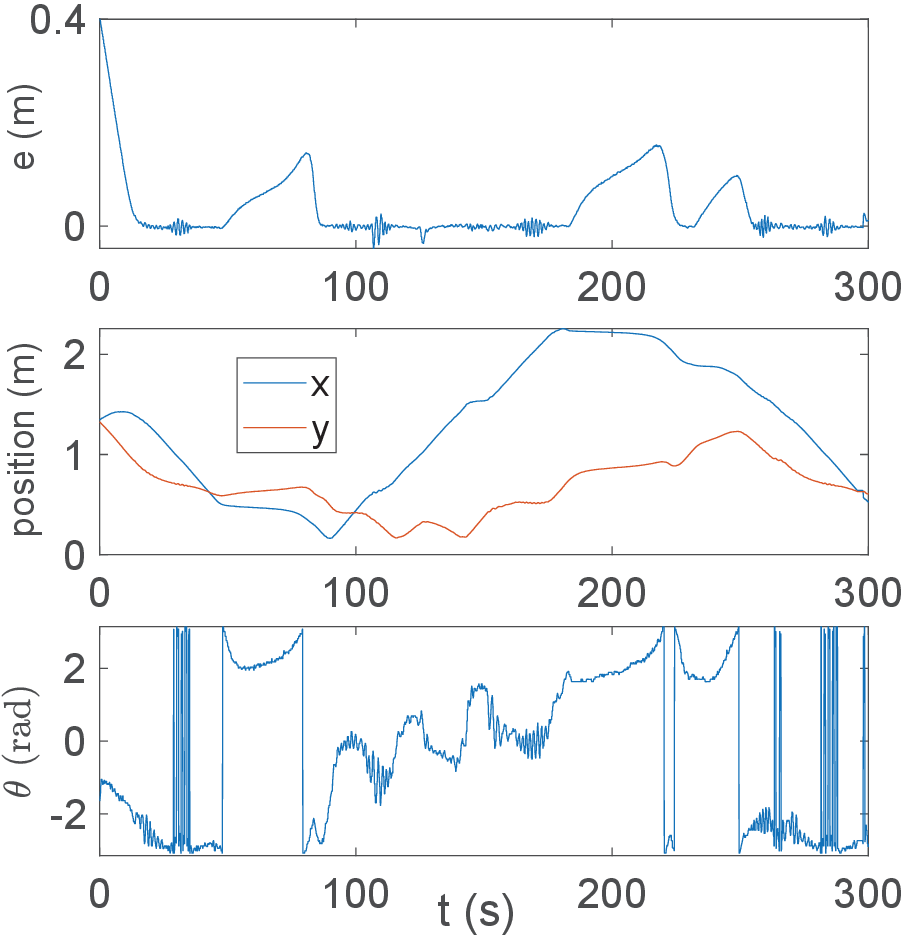}
		\label{case2position}}
	\subfigure[Robot velocities.]{
		\includegraphics[width=4.2cm]{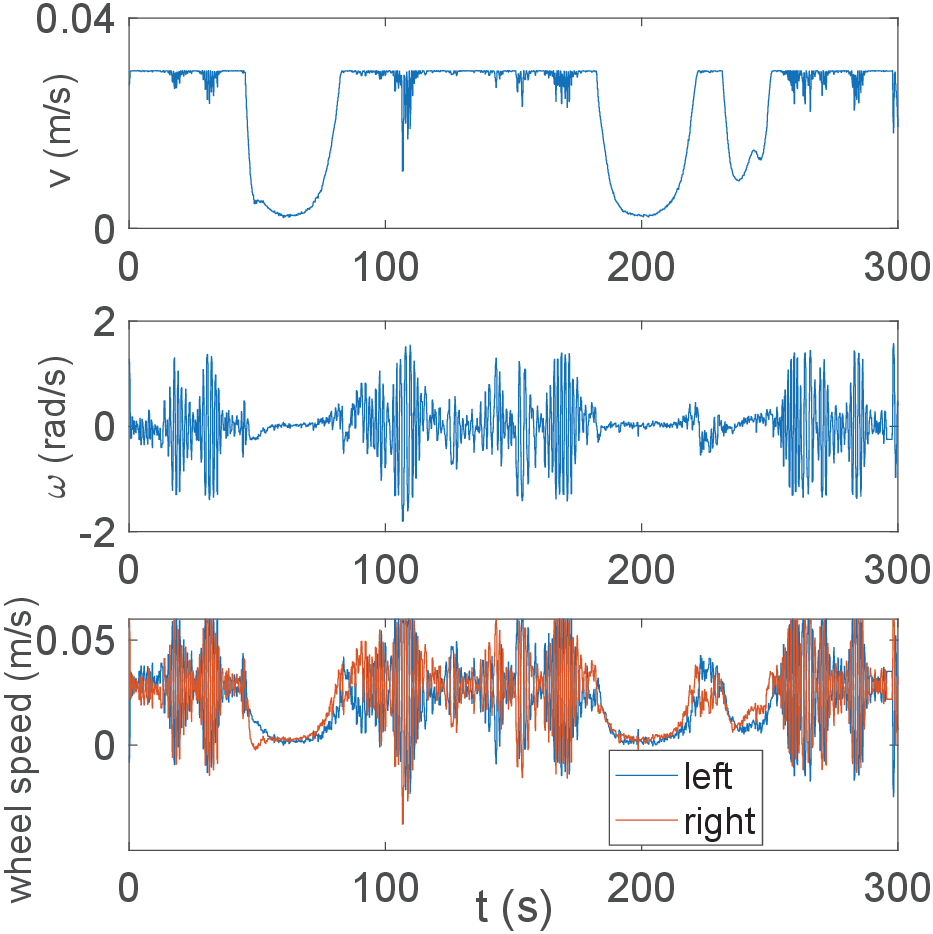}
		\label{case2velocity}}
	\caption{Time histories in experiment 2.}
	\label{exp2time}
\end{figure}

In the second experiment, the robot starts from $[1.3,1.3]^\top$ m with an orientation of $92$ deg, and three obstacles are set on the boundary. As the boundary is decomposed into two segments, the VF switches from one to another at $x=1$ m, which is consistent with Fig. \ref{point}. Control parameters are taken as $k=15$, $v_d=0.03$ m/s, $v_m=0.06$ m/s, and results are shown in Figs. \ref{exp2traj}-\ref{exp2time}. The green curve shows the trajectory of the robot, which sees unwanted oscillations. We found that it was mainly caused by the time delay of camera signal transmission to obtain the robot states. 

\begin{figure*}[ht]
	\centering
	\subfigure{
		\includegraphics[width=5.4cm]{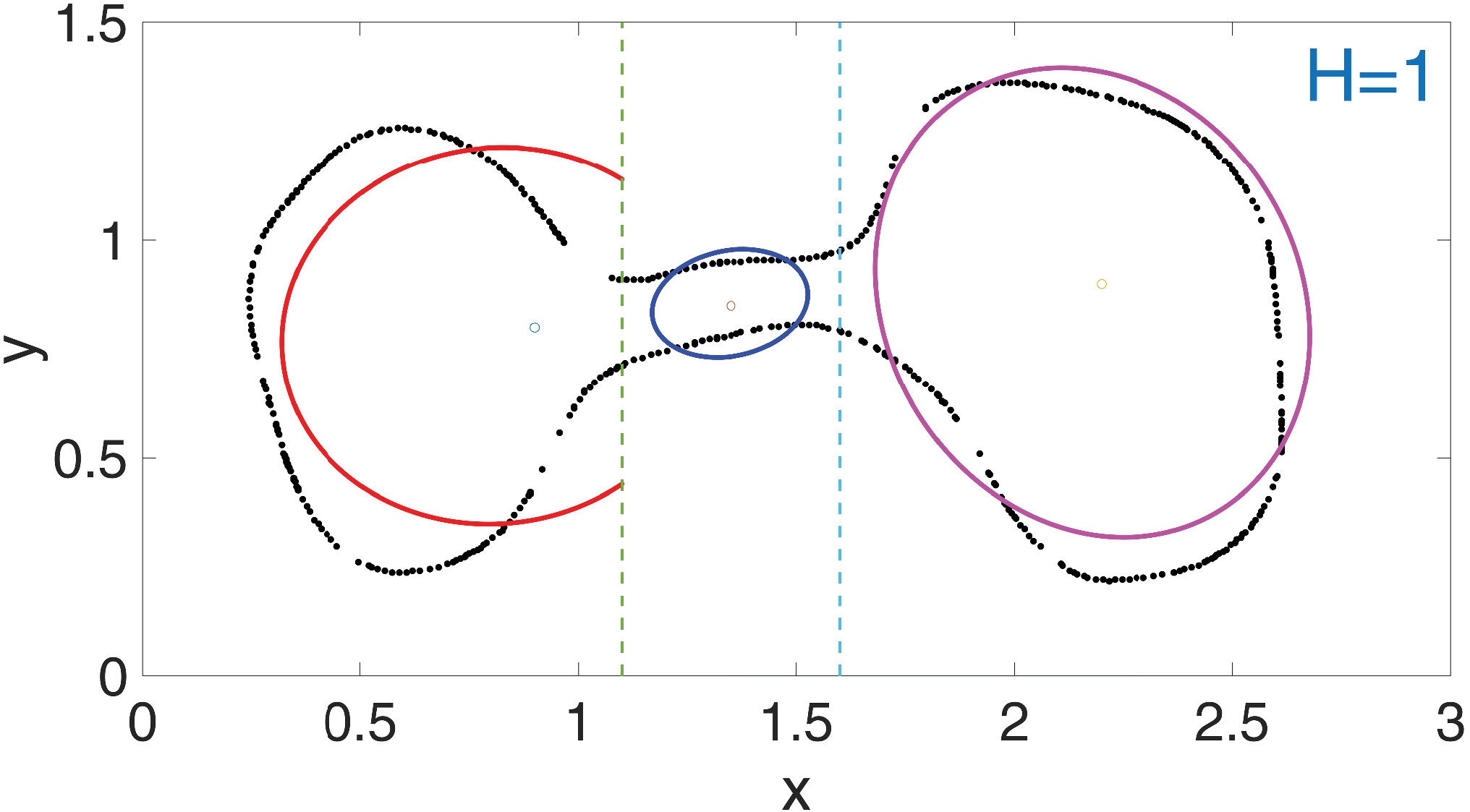}
		\label{h=1}}
	\subfigure{
		\includegraphics[width=5.4cm]{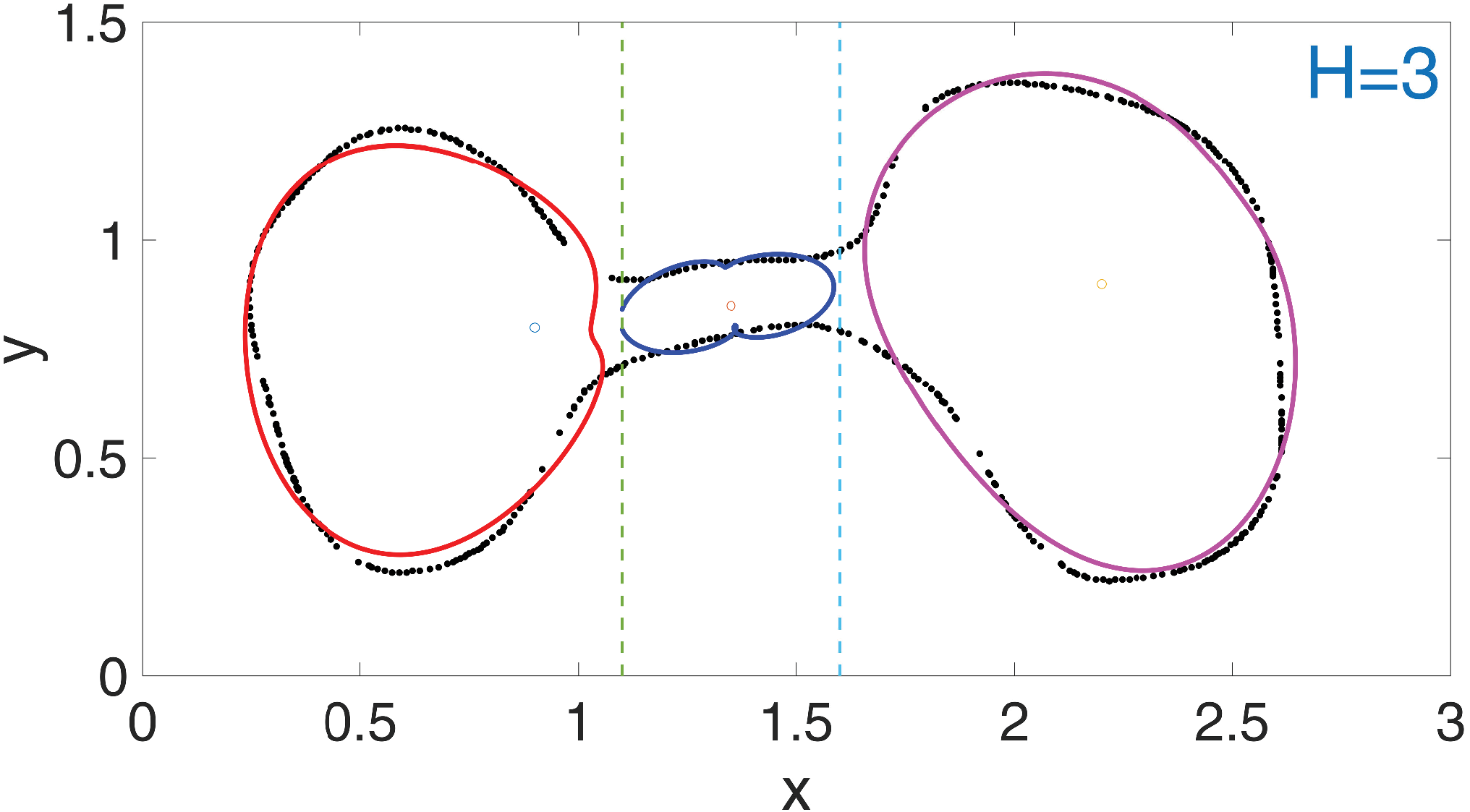}
		\label{h=3}}
	\subfigure{
		\includegraphics[width=5.4cm]{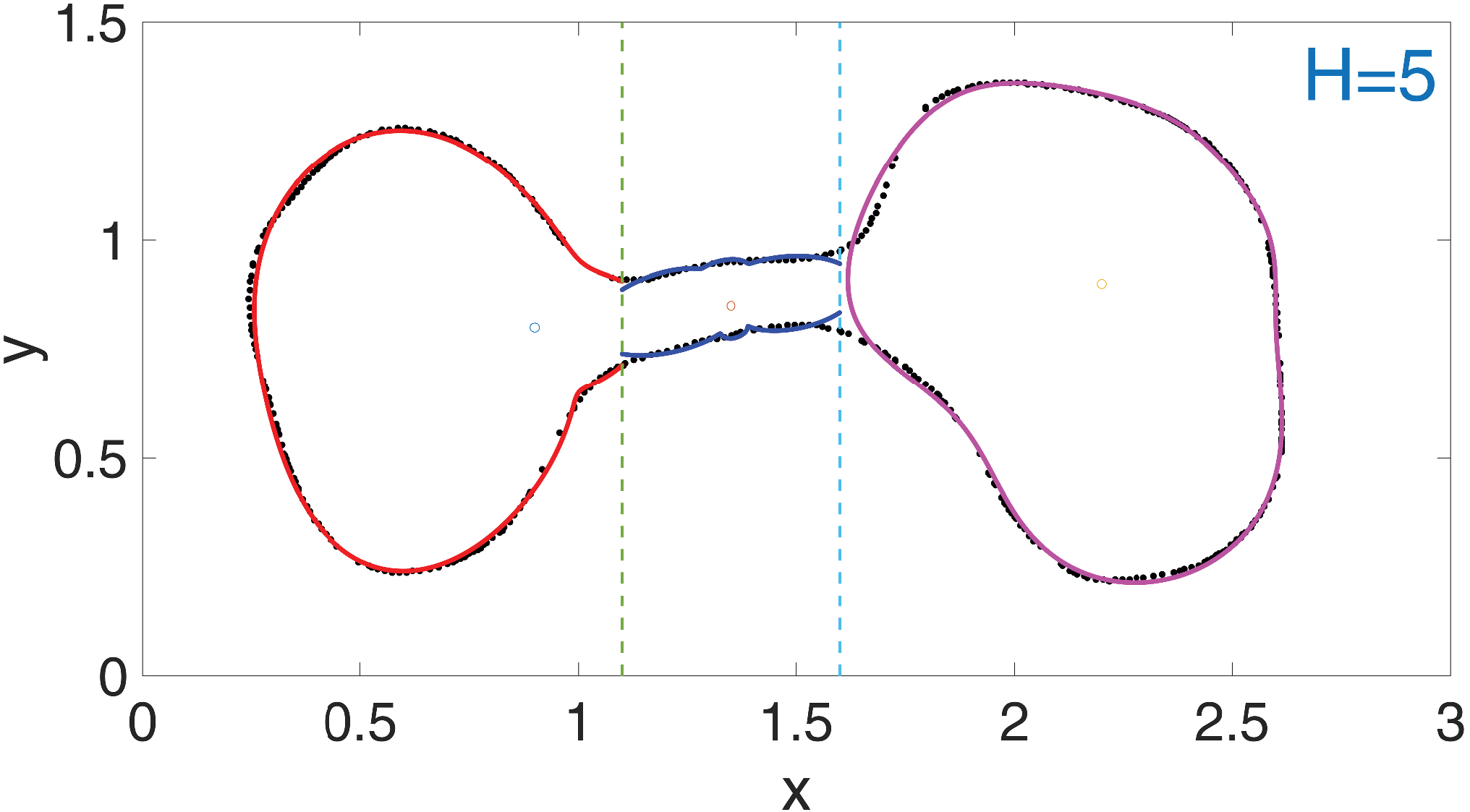}
		\label{h=5}}
	\subfigure{
		\includegraphics[width=5.4cm]{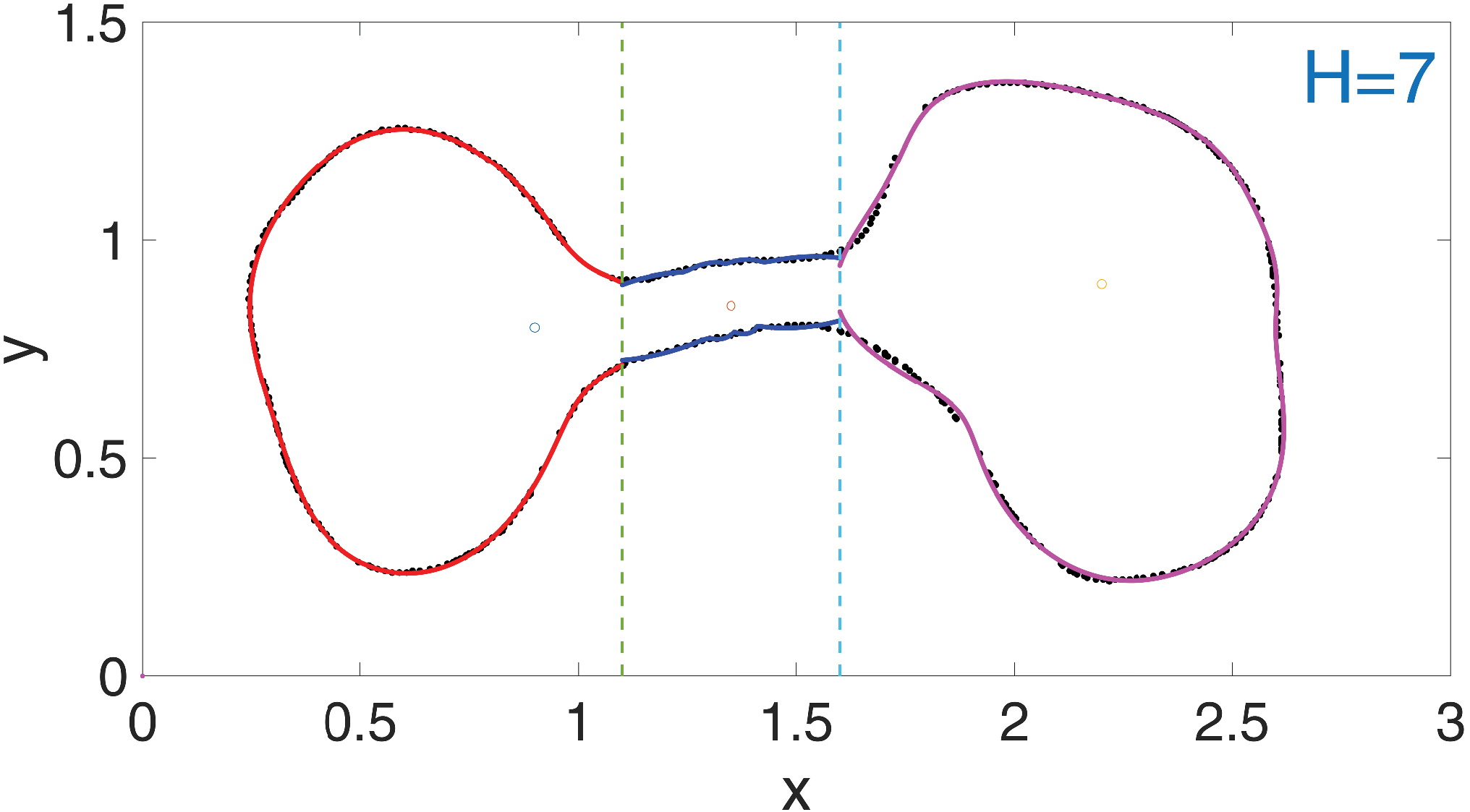}
		\label{h=7}}
	\subfigure{
		\includegraphics[width=5.4cm]{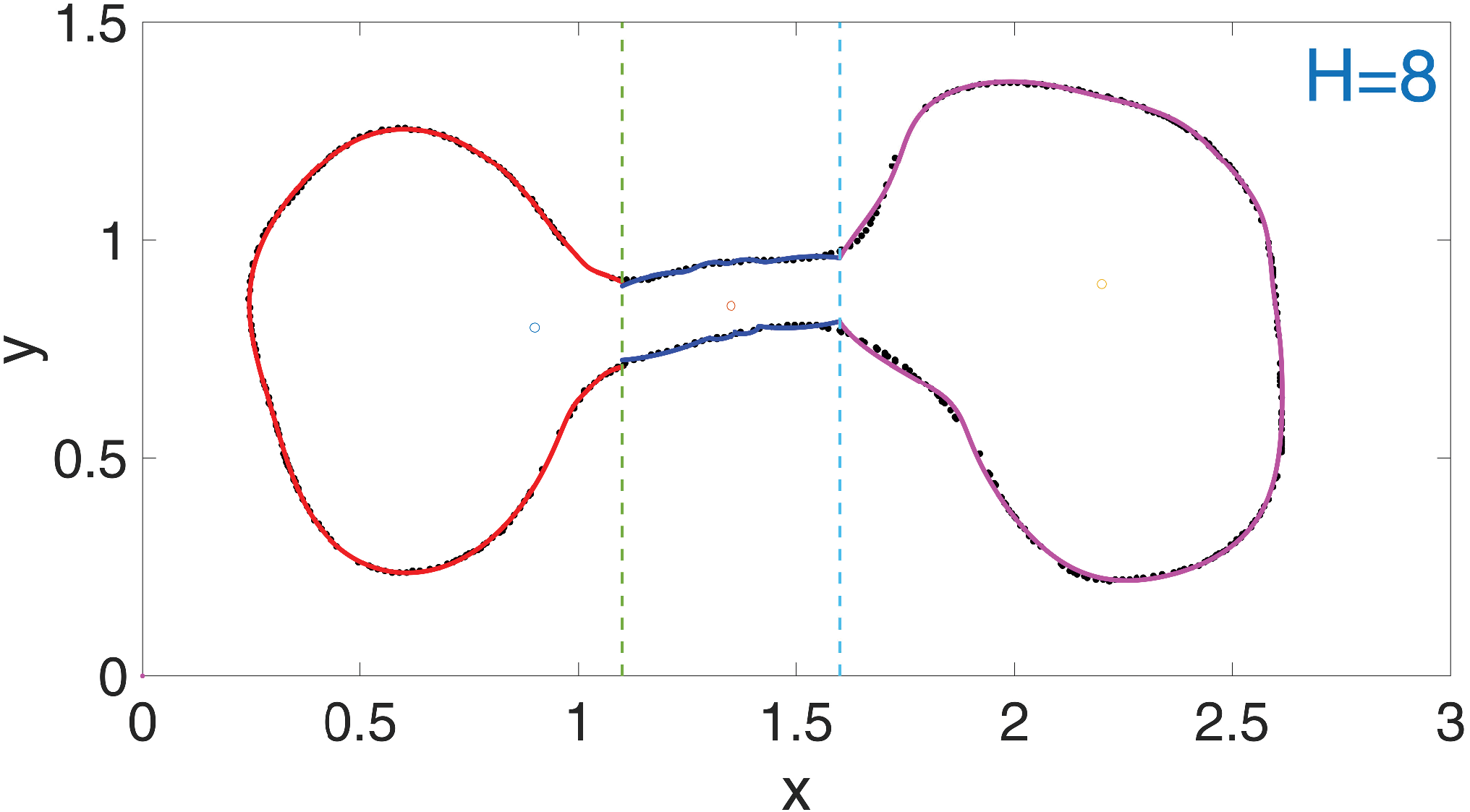}
		\label{h=8}}
	\subfigure{
		\includegraphics[width=5.4cm]{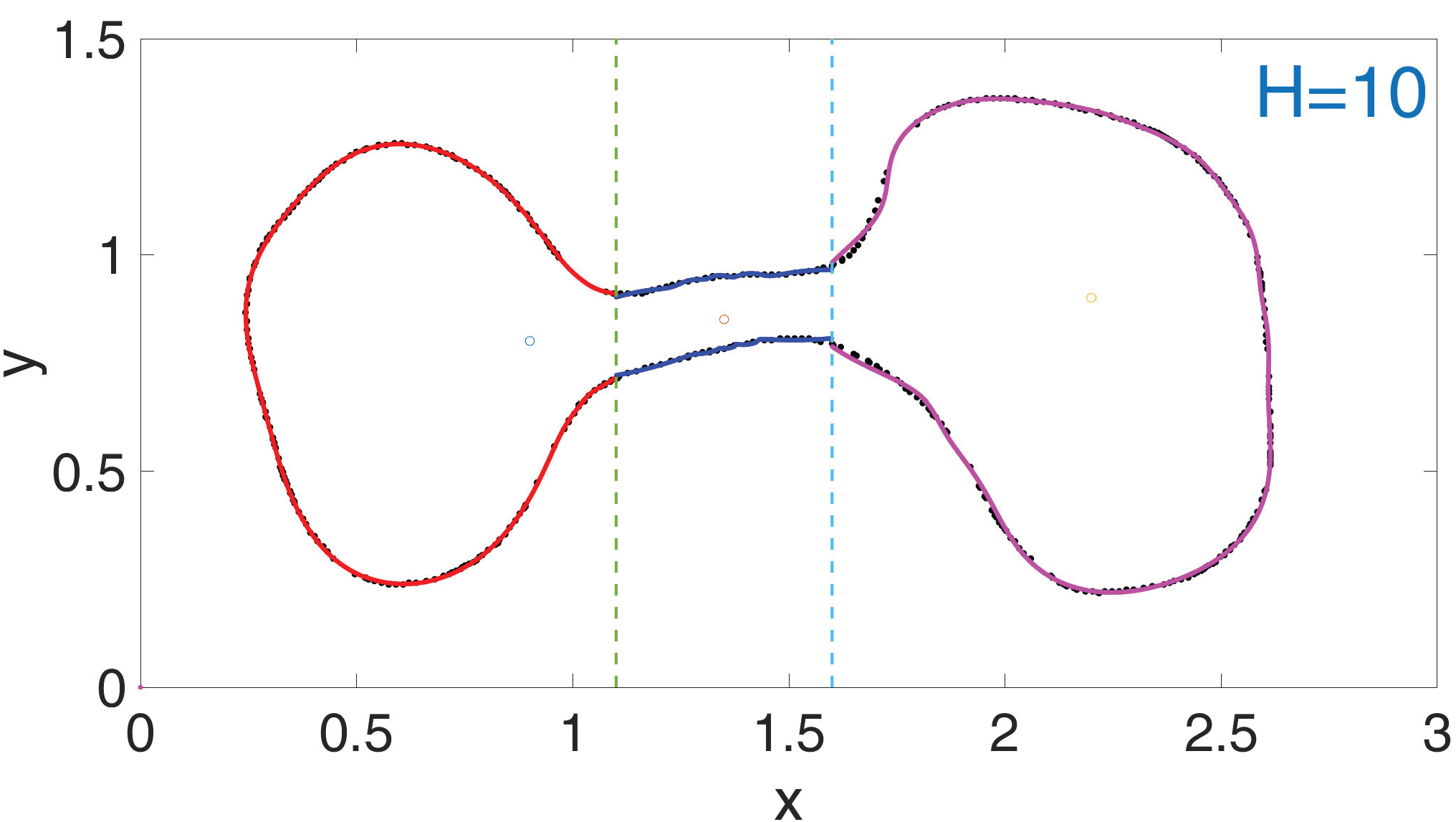}
		\label{h=10}}
	\caption{Experiment 3 boundary approximation. The curve fitting precision increases as more harmonics (larger $H$) are considered.}
	\label{exp3fitting}
\end{figure*}

\begin{figure*}[ht]
	\centering
	\subfigure{
		\includegraphics[height=4.2cm]{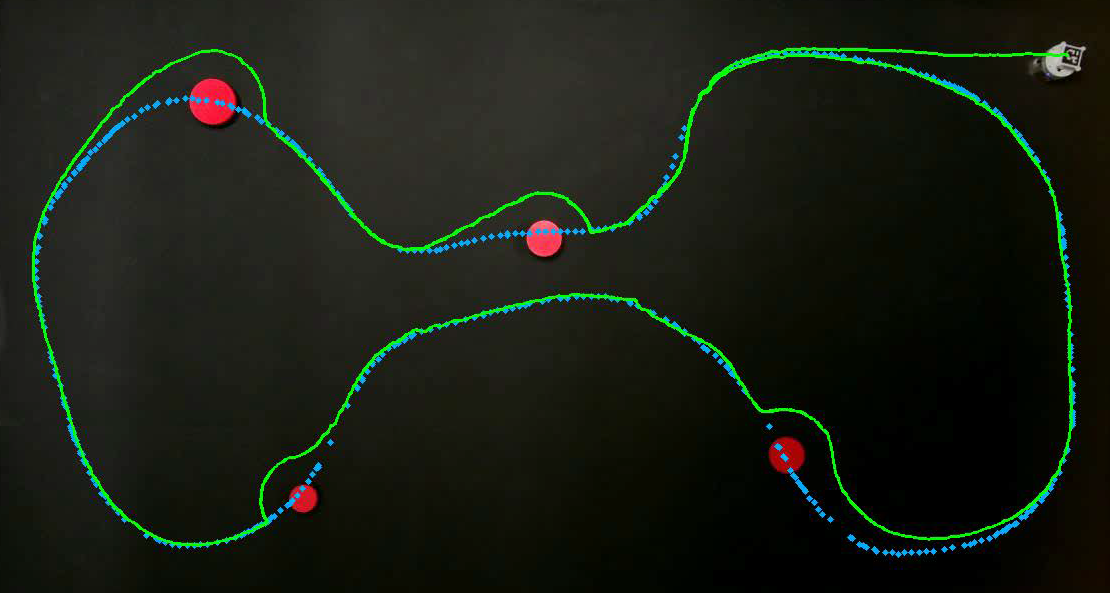}
		\label{exp3traj}}
	\subfigure{
		\includegraphics[height=4.2cm]{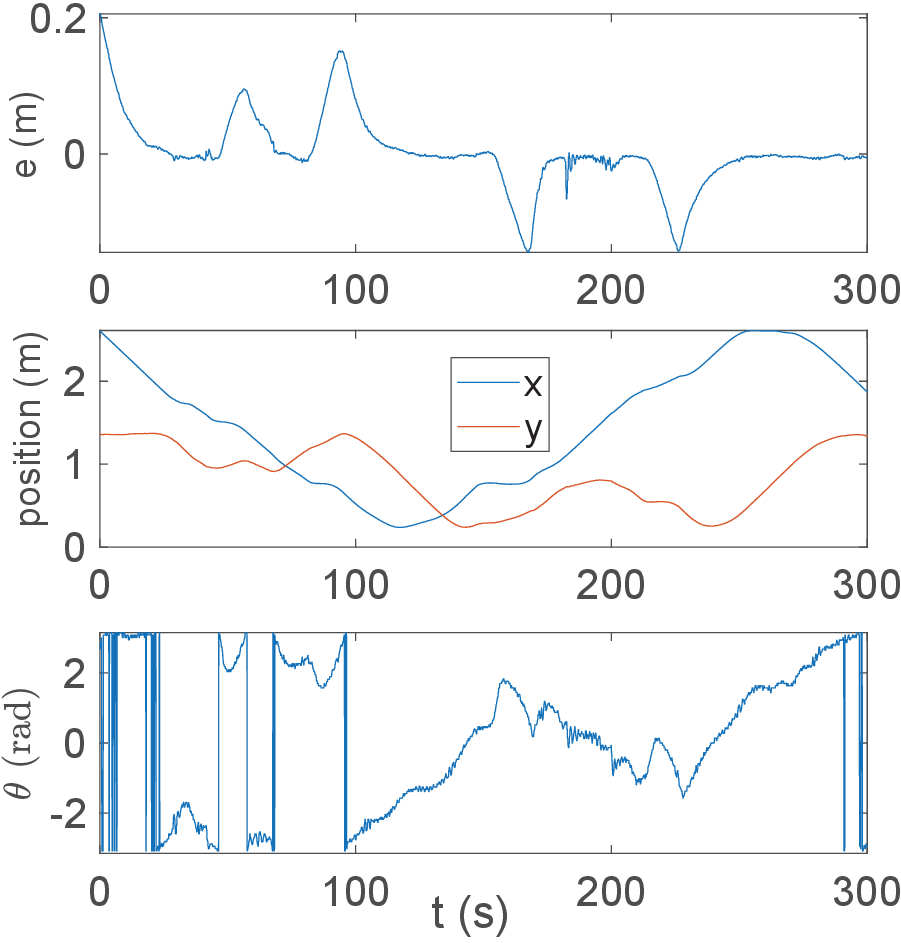}
		\label{case3position}}
	\subfigure{
		\includegraphics[height=4.2cm]{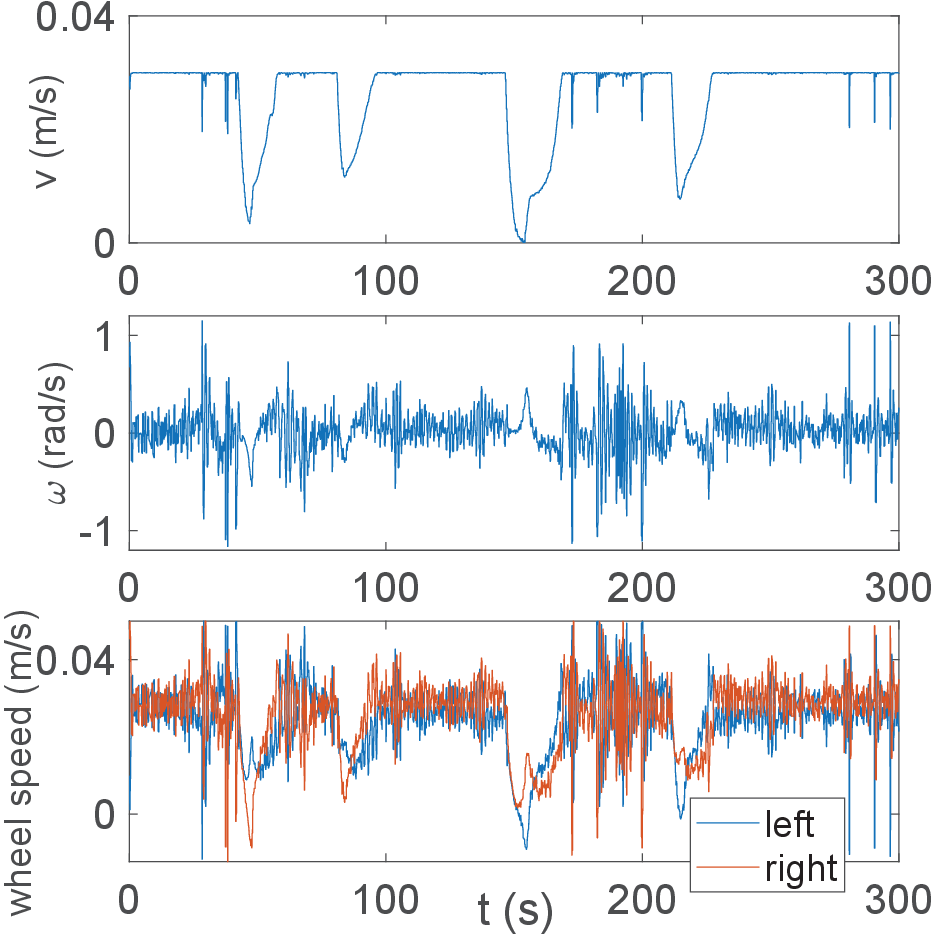}
		\label{case3velocity}}
	\subfigure{
		\includegraphics[width=17cm]{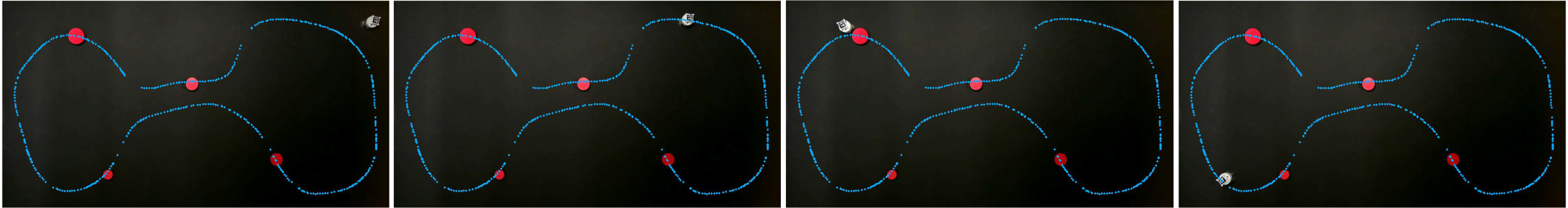}
		\label{exp3shot}}	
	\caption{Experiment 3 results. Top: Robot trajectory and time histories. Bottom: Snapshots during the experiment.}
	\label{exp3}
\end{figure*}

Besides, in the third experiment, a more complex boundary is considered, which needs to be approximated on three different segments. We separate the region into $x<1.1$ m, $1.1$ m$<x<1.6$ m and $x>1.6$ m, respectively, as shown in Fig. \ref{exp3fitting}. It is seen that the boundary is fitted with increasing precision as more harmonics are taken into consideration, and $H=10$ is used for boundary representation. Fig. \ref{exp3} presents the results, wherein the robot gets closer to the boundary and then encircles it, successfully bypassing four obstacles.


Furthermore, we conduct multiple experiments with different initial positions and obstacle placements for these three experiments, and more details are provided in the supplementary video.

\section{Conclusion}
This article presented a control scheme for mobile robots to encircle general 2-D boundaries, which can be applied to perform some practical tasks such as cleaning chemical spills and environment monitoring.
A Fourier-based boundary fitting method is developed to approximate its parametric equation. The advantage is that, only sampled points are used for boundary characterization, with no need for other equations in advance. By decomposing the boundary into several star-shaped sets, our proposed method can characterize general boundaries with any shape. 
In addition, a VF was designed for guiding the encirclement of the parameterized boundary, and the synthesized controller further guarantees satisfaction of obstacle avoidance and input saturation constraints.
Simulation and experimental results show the effectiveness of our proposed algorithm.
Future works may focus on extending the proposed method to multi-robot systems, and considering dynamic obstacles.

\ifCLASSOPTIONcaptionsoff
  \newpage
\fi

\bibliographystyle{IEEEtran}
\bibliography{main}

\end{document}